\documentclass[letterpaper,twocolumn,10pt,table]{article}
\usepackage{usenix-2020-09}

\usepackage{tikz}
\usepackage{amsmath}

\usepackage{filecontents}

\usepackage{multirow}
\usepackage{xcolor}
\usepackage{xspace}
\usepackage{xurl}

\usepackage{comment}
\usepackage{csquotes}

\usepackage[vskip=0pt]{quoting}

\usepackage{booktabs}
\usepackage{dcolumn}

\usepackage{amsmath}

\usepackage{adjustbox}

\usepackage{tcolorbox}
\usepackage{amssymb}

\usepackage{array}
\usepackage{enumitem}

\usepackage{changepage}

\usepackage{etoolbox}
\usepackage{pgf}
\usepackage{highlight}
\usepackage{scrextend}

\usepackage{etoolbox}

\excludecomment{usenix}

\includecomment{arxiv}

\newif\ifremovemarkers
\removemarkersfalse 
\ifremovemarkers
\newcommand{\omer}[1]{{}}
\newcommand{\nina}[1]{{}}
\newcommand{\sai}[1]{{}}
\newcommand{\michelle}[1]{{}}
\newcommand{\animesh}[1]{{}}
\newcommand{\benoit}[1]{{}}
\newcommand{\hamza}[1]{{}}

\else
\newcommand{\omer}[1]{{#1}}
\newcommand{\nina}[1]{{#1}}
\newcommand{\sai}[1]{{#1}}
\newcommand{\michelle}[1]{{\color{blue}#1}}
\newcommand{\animesh}[1]{{\color{blue}#1}}
\newcommand{\benoit}[1]{{\color{blue}#1}}
\newcommand{\hamza}[1]{{\color{blue}#1}}

\fi

\renewcommand{\sectionautorefname}{\S\kern-0.2em}
\renewcommand{\subsectionautorefname}{\S\kern-0.2em}
\renewcommand{\subsubsectionautorefname}{\S\kern-0.2em}

\newcommand{\turkey}{Türkiye\xspace}

\newcommand{\Theme}{Theme\xspace}
\newcommand{\Themes}{Themes\xspace}

\newcommand{\theme}{theme\xspace}
\newcommand{\themes}{themes\xspace}
\newcommand{\themed}{themed\xspace}

\newcommand{\Issues}{Issues\xspace}

\newcommand{\issues}{issues\xspace}

\newcommand{\tgtwo}{app type\xspace}
\newcommand{\tgtwohyphen}{app-type\xspace}

\newcommand{\tgtwos}{app types\xspace}

\newcommand{\TgTwos}{App Types\xspace}

\renewcommand{\sp}{security \& privacy\xspace}

\newcommand{\ppr}{PPR\xspace}
\newcommand{\pprc}[1]{$PPR_{#1}$\xspace}
\newcommand{\pprall}{$PPR_{(all)}$}\xspace
\newcommand{\ptpr}{PTPR\xspace}
\newcommand{\ptprc}[1]{$PTPR_{#1}$\xspace}

\newcommand{\dev}{developer\xspace}

\newcommand{\googleplay}{developer-specified\xspace}

\newcommand{\approach}{data-analysis pipeline\xspace}

\newcounter{findingcounter}
\newcommand{\takeaways}{ \smallskip\noindent\textbf{{\bf Finding \arabic{findingcounter}:}\phantom{xx}}\addtocounter{findingcounter}{1}}

\newcolumntype{R}[1]{>{\raggedleft\arraybackslash}p{#1}}
\newcolumntype{L}[1]{>{\raggedright\arraybackslash}p{#1}}

\newcommand{\g}[1]{\gradientcelld{#1}{0.65}{0.85}{1}{red}{white}{green}{70}}

\renewenvironment{quotation}
       {\list{}{\listparindent=2pt
                \itemindent    \listparindent
                \leftmargin=7pt
                \rightmargin=7pt
                \topsep=2pt
                }%
        \item\relax}
       {\endlist}

\begin{document}

\date{}

\title{\Large \bf A Decade of Privacy-Relevant Android App Reviews: Large Scale Trends}

\author{
{\rm Omer Akgul}\\
University of Maryland
\and
{\rm Sai Teja Peddinti}\\
Google
\and
{\rm Nina Taft}\\
Google
\and
{\rm Michelle L. Mazurek}\\
University of Maryland
\and 
{\rm Hamza Harkous}\\
Google
\and
{\rm Animesh Srivastava}\\
Google
\and
{\rm Benoit Seguin}\\
Google
} 

\date{%
    $^1$Organization 1\\%
    $^2$Organization 2\\[2ex]%
}

\maketitle

\begin{abstract}

    We present an analysis of 12 million instances of privacy-relevant reviews publicly visible on the Google Play Store that span a 10 year period. By leveraging state of the art NLP techniques, we examine what users have been writing about privacy along multiple dimensions: time, countries, app types, diverse privacy topics, and even across a spectrum of emotions. We find consistent growth of privacy-relevant reviews, and explore topics that are trending (such as Data Deletion and Data Theft), as well as those on the decline (such as \omer{privacy-relevant reviews on sensitive permissions}). We find that although privacy reviews come from more than 200 countries, 33 countries provide 90\% of privacy reviews. We conduct a comparison across countries by examining the distribution of privacy topics a country's users write about, and find that geographic proximity is not a reliable indicator that nearby countries have similar privacy perspectives. We uncover some countries with unique patterns and explore those herein. Surprisingly, we uncover that it is not uncommon for reviews that discuss privacy to be positive (32\%); many users express pleasure about privacy features within apps or privacy-focused apps. We also uncover some unexpected behaviors, such as the use of reviews to deliver privacy disclaimers to developers.
    Finally, we demonstrate
    the value of analyzing app reviews with our approach as a complement to existing methods for understanding users' perspectives about privacy.
    
\end{abstract}

\section{Introduction}
User perspectives are commonly measured through user studies (e.g., surveys, interviews, lab studies), which can provide rich data to answer focused research questions. Unfortunately, such studies do not scale beyond thousands of users, and the resulting measurements are heavily bound to the 
environment the studies were conducted in: 
user opinions may vary over time, between regions, and across different app types. Rerunning user studies to understand these differences can be prohibitively costly, both monetarily and time-wise. 

In this work, we present an alternate and complementary analysis approach, with a different set of research tradeoffs. Specifically, we trade the power of surveys in having the same (targeted) questions answered by many participants for the power of ecologically valid, large-scale analysis in uncovering unanticipated insights from open-ended reviews. We believe these insights can, in turn, motivate future investigations of user-centered privacy, including survey-based studies.

Getting input about users' perspectives on privacy issues from millions of user reviews from hundreds of countries has not been feasible until recently. Advances in natural language processing (NLP) and large language models (LLMs) permit complex analysis of enormous text corpora. Here, we leverage these advances to investigate users' privacy opinions and concerns from a novel perspective: 
we study 12.3M (million) privacy-related reviews, extracted from $\sim$2B (billion) public reviews on Google Play spanning 10 years  (Jan. 2013--Feb. 2023). These privacy reviews come from more than 200 countries or regions, 25 languages, and 160K (thousand) apps that span every Play app category.

To analyze this extensive dataset, we leverage and extend NLP techniques recently introduced in Hark~\cite{harkous20222hark} to automatically extract all reviews 
that discuss a privacy topic, assign fine-grained issue tags to each review, aggregate related issues into larger thematic clusters, and classify the ``emotions'' expressed in  these reviews.
The resulting dataset of roughly 12.3M privacy-related reviews likely constitutes the largest body of privacy 
feedback ever evaluated at this granularity.
Using this dataset, we address the following research questions: 
\begin{itemize}[itemsep=-0.2em]
    \item \textbf{RQ1:} Which privacy issues do users raise and discuss in app reviews?
    \item \textbf{RQ2:} How have these privacy issues evolved over time? 
    \item \textbf{RQ3:} How do privacy issues vary across countries? 
    \item \textbf{RQ4:} Which types of apps have privacy reviews with strongly negative or strongly positive emotions?
    \item \textbf{RQ5:} How do reviews as a source of understanding privacy perspectives complement prior work?
\end{itemize}

We find that privacy reviews have grown 
steadily over 10 years, both in terms of absolute volume (a $4.7$x increase) and when normalized for review volume (9\% biannual growth). 
We find that \themes such as \texttt{Data Deletion} are growing in importance, while reviews relating to \texttt{Excessive \omer{[Privacy-Relevant]} Permissions} (a popular research topic, e.g., \cite{peddinti2019permissions,cao2021large, felt2011android,wijesekera2015android}) have seen a significant decline.

Our broad overview across the globe 
also shows that geographic proximity is a 
weak indicator of whether nearby countries
discuss similar privacy issues.  We find that the countries that contribute the largest volume of privacy related reviews tend to be countries with large populations, and not necessarily countries that drive privacy regulation (such as the EU). We uncover a handful of countries (typically understudied) that discuss unique distributions of privacy topics. For example, in \turkey, we find a significant number of reviews using almost the same quasi-legal disclaimers---implying users might be under the mistaken assumption that this bolsters privacy protection.

The app types whose privacy reviews exhibit the most strongly negative emotions (e.g. anger, annoyance, fear) are social media apps, parental control and child monitoring apps, as well as simulation games that mimic users facial and vocal expressions (leading to anxiety about surveillance). On the other hand, reviews for security and privacy apps commonly express positive emotions.

Finally, we place our results in context of related research, adding context to prior findings and identifying new areas for further study. For example, we expand on previously documented privacy concerns related to loan apps~\cite{munyendo2022kenya}, finding that these concerns arise in a number countries beyond the original one identified. We also 
observe strongly positive reviews of many apps that claim to secure or hide content on phones, especially in the context of multi-user devices~\cite{sambasivan2018privacy}, raising a number of questions for future research. We demonstrate the utility of automatically distilling user feedback at scale, as a complement to other methods of understanding users' opinions and concerns.

\section{Background and Related Work}
\label{sec:related-work}

Interviews and surveys are the most 
common method of measuring privacy attitudes. They have been broadly used to understand users' mental models of security \& privacy 
tools~\cite{krombholz2019if, akgul2021e2ee, binkhorst2022vpn}, to study privacy preferences for smartphone app permissions \cite{cao2021large,reyes2018won}, to measure privacy concerns with IoT and sensors~\cite{rader2022sensors, malkin2019privacy, zhao2022visualimpairmentcamera, ur2012smart}, and used in methods studies~\cite{tang2022privacy, abrokwa2021iosvandroid, redmiles2019generalize}---to name only a few in this vast field. While these methods are sound and routinely used, researchers also acknowledge limitations arising from social desirability, acquiescence, and demand biases. While behavioral measurements can circumvent some of these biases~\cite{bitton2020evaluating, egelman2016behavior}, they depend on inferring what users think rather than being able to document exactly what users' attitudes are. Moreover, recent work presents evidence that problems with validity of privacy constructs (built from surveys) may be widespread~\cite{conago2022privacyscale}.
Similar methods are used to measure developers' 
mental models of privacy threats~\cite{mhaidli2019devads}, their understanding of app 
privacy~\cite{mhaidli2019devads}, and their responsiveness to privacy 
nudges~\cite{tahaei2021devads}. Studying developers themselves is out of scope of this work.

A key difference in our work is our method: namely, 
recently proposed LLM techniques, specific to the app 
reviews context~\cite{harkous20222hark}, enabling us 
to analyze inputs from millions of users. 
This method, complementary to those above, offers a different set of tradeoffs. While interviews allow for in-depth questioning of what users think, they typically do not scale beyond a few tens of users. Surveys can scale to thousands (but not millions) of users and enable the development of verifiable privacy scales, but are very focused; a researcher cannot learn about an issue that wasn't posed in the questionnaire. Further, recruitment across multiple geographic regions is expensive and challenging, making such studies often costly to repeat.
In contrast, our approach does not support controlled experiments or hypothetical questions about potential new designs. Still, it does allow for increased ecological validity and much larger and broader samples.

In this paper, we explore what users around the globe write about privacy in Android app reviews. A few works have explored app reviews~\cite{nguyen2019short,Nema2022-ICSE,harkous20222hark}; however, they extracted a limited set of privacy reviews because their privacy classifiers relied on keywords \cite{nguyen2019short}, limited heuristics for data sampling \cite{Nema2022-ICSE}, or a limited privacy taxonomy \cite{harkous20222hark}. These works also only included English language reviews, whereas we include 25 languages. Our set of privacy-related reviews is 1000x, 20x and 2x larger (respectively) than these earlier studies.

Many cross-country studies focus on specific aspects of privacy, such as android 
permissions~\cite{cao2021large}, social networks~\cite{wang2011uscninprivacy}, phone locking 
behavior~\cite{harbach2016keep}, incident response~\cite{redmiles2019should}, or how much users are willing to pay for privacy for specific types of data~\cite{prince2022much}.
Beyond these focused studies, some more general cross-country studies attempt to understand the influence of factors such as culture or country on privacy attitudes or privacy preferences~\cite{anaraky2021privacyculture,cho2009multinational}. Some studies do show differences between non-Western and Western countries in terms of misconceptions around privacy~\cite{herbert2022world} practices.

A recent survey~\cite{hasegawa2023survey} notes limited geographic diversity in usable security \& privacy
research, with participants primarily from Western, educated, industrialized, rich, and democratic (WEIRD) societies. While this may naturally occur due to geographic and linguistic barriers, our approach---using text from > 200 countries/regions and translations for 24 languages---offers an alternative. 
Most of the multinational surveys above include 3-7 countries; a few include 10-20. To our knowledge, our work is the first to report on data from more than 200 countries, with the top-50 explicitly compared.

\section{Data Analysis Pipeline}

In this section, we describe our \approach that re-uses several lessons from the Hark system~\cite{harkous20222hark}, highlighting the modifications we made to fit the purpose of this study. We also discuss the 
resulting dataset and our analysis approach.

\subsection{Text Analysis Pipeline}
\label{sec:hark_modifications}

\begin{figure}[tb!]
    \centering
    \includegraphics[width=.48\textwidth]{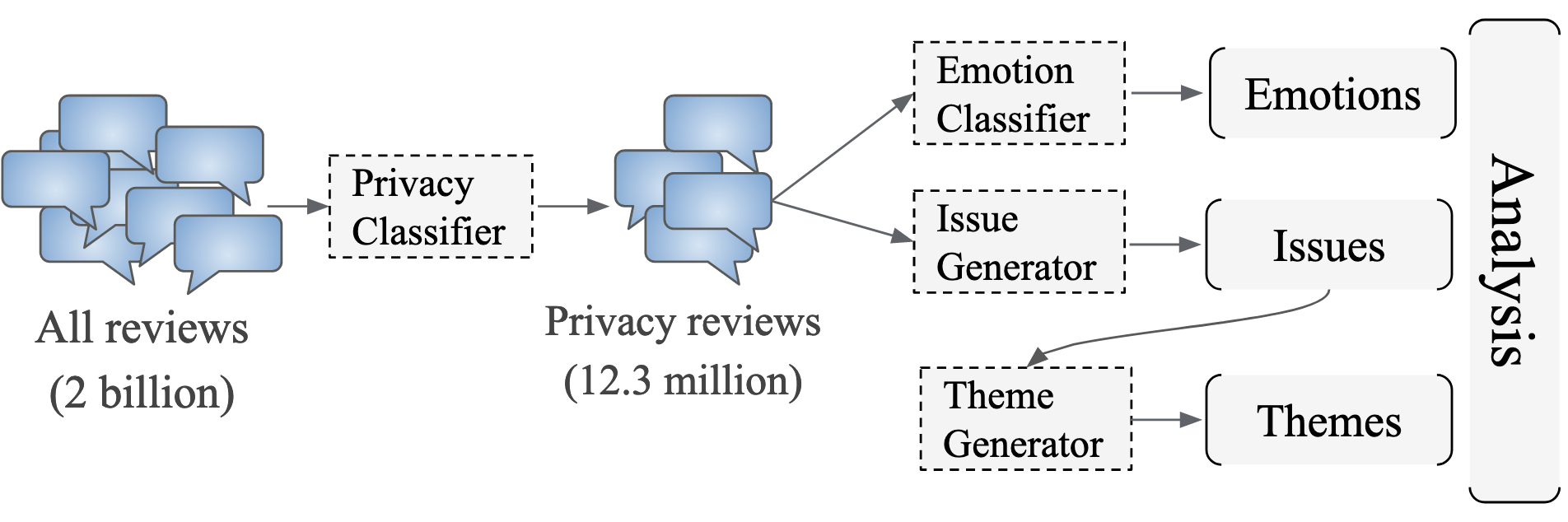}
    \caption{Our analysis pipeline overview, adapted from~\cite{harkous20222hark}.}
    \label{fig:hark_pipeline}
\end{figure}

We build our analysis pipeline to leverage and extend the components in Hark, an end-to-end system for retrieval and analysis of privacy-related feedback leveraging state-of-the-art techniques in NLP~\cite{harkous20222hark}. An overview of our \approach is in~\autoref{fig:hark_pipeline}. First, the \textit{privacy classifier} identifies the privacy-related feedback from unstructured text. The \textit{issue generation} model takes in this privacy feedback and dynamically generates meaningful, fine-grained issues (covering both known and newly emerging issues) describing the privacy aspects discussed within each review text. The \textit{theme creation} component groups these issues into thematic clusters and assigns a succinct title to each. Our \approach also includes an \textit{emotion} classifier that dissects each review's text across 28 emotions (e.g., anger, fear, joy and confusion). 

We obtained models and 
training data from 
Hark~\cite{harkous20222hark} and re-purposed most of the pipeline, with modifications to the privacy classifier
to improve the breadth of topics identified.
First, we expand the privacy taxonomy from the 
original paper (25 concepts) to include 89 
privacy-relevant 
concepts\begin{usenix}
.\footnote{\sai{See the extended paper for the expanded privacy taxonomy~\cite{akgul2024reviews}.}} 
\end{usenix}
\begin{arxiv}
  (see \autoref{app:taxonomy}).
\end{arxiv}
The prior taxonomy already included concepts from multiple known taxonomies. However, when we asked five privacy experts (with at least 5 years of academic/industry experience in privacy research and development) to manually examine 
the original taxonomy~\cite{harkous20222hark}, they observed many missing topics. For example, we added Data Hiding, Opt-out and Location concepts, and added specificity to the Anonymity and Personal Data types. 

Second, we rely on these same experts, instead of crowd-sourcing, to manually annotate a new training dataset for the privacy classifier, based on the expanded taxonomy and following the same \textit{Natural Language Inference} approach and heuristics as in~\cite{harkous20222hark}. To best use the experts' time and generate high-quality training data without requiring multiple annotations per review, we used two labeling rounds. In the first round, we described the taxonomy to the experts and each annotator independently labeled a subset of reviews. Following the principles of \textit{Active Learning}, we trained a classifier on the annotated reviews from the first round. The active-learning classifier is composed of Sentence-T5 frozen embeddings~\cite{ni-etal-2022-sentence}, a dense layer, and a binary classification head. The experts discussed cases where the model produced a different label from the expert as well as cases whose classification probabilities had high entropy. In the second round, all experts discussed these misclassifications or low-confidence predictions, thereby focusing their efforts on challenging examples. Overall, we generated a labeled dataset of 4.3K (nearly equal split of 2K \textit{privacy} and 2.3K \textit{not-privacy}) reviews, which we split into training, validation, and test sets. We fine-tuned T5-11B~\cite{raffel_t5} on these datasets and use it as our privacy classifier.

\paragraph{\omer{Model evaluation:}} Our classifier 
has a 0.95 ROC AUC, 
87\% precision, and 86\% recall on our 
new diversified test set,\footnote{We set a 0.8 prediction score threshold for best accuracy and verified results qualitatively.} 
and a similar ROC AUC of 0.88 (vs 0.92) 
on the original Hark test set~\cite{harkous20222hark}.
In comparison, the privacy classifier from~\cite{harkous20222hark}
has 0.87 ROC AUC, 
89\% precision, and 51\% recall on our new test set,
demonstrating that our classifier performs better 
at capturing the diversity of our taxonomy. \nina{(For additional experiments with a range of model architectures see \autoref{app:classifier_baselines}.)}
\sai{We also performed qualitative assessment of
all 50 false positives/negatives of our classifier, noticing primarily issues with reviews that were short ($\leq$ 10 words) and ambiguous (missing context, causing multiple possible interpretations). For example, 
a short review ``\textit{Taking too much data}'' was labeled as `privacy,' when the privacy expert (based on other reviews seen) interpreted it to be a complaint about the app consuming the limited mobile data bandwidth and annotated it as `not-privacy'; whereas an ambiguous review ``\textit{msgs have been deleted, but for some reason they remain in place}'' was labeled as `not-privacy' (could be seen as a bug in app functionality), when the privacy expert interpreted it as data deletion control not working as expected (a privacy concern). 13 of the 50 false positives/negatives were short; and the rest (though longer) were still ambiguous. Despite missing these short and ambiguous cases, our classifier had good performance.}

Similar to \cite{harkous20222hark}, for ease of representation, we consolidated the 28 classifier-generated emotions into 8 emotions groups (plus a neutral option) based on Ekman’s emotions taxonomy~\cite{ekman1992argument} and using Demszky et al.'s~\cite{demszky2020goemotions} grouping criteria.

\subsection{Data Description}
\label{sec:data-description}

We obtained access to reviews directly from the Google Play team. 
Our initial dataset consists of all $\sim$2B publicly visible 
Google Play app reviews for apps with $>$10K installs, 
spanning 10 years from Jan 2013--Feb 2023. The dataset was already anonymized (no user identifying information) and sanitized (any detected fake/spam reviews were removed).  
Each review is associated with the review text, its language, country, submission time, star rating, the corresponding app's package name, the app’s \googleplay category information and 
(if available) finer-grained \textit{\tgtwohyphen} 
information (such as `Rideshare \& Taxis' 
for a rideshare app, which belongs 
to the `Maps \& Navigation' \googleplay 
category). Note that we do not generally use 
\googleplay categories as they are too broad~\cite{peddinti2019permissions}. Instead we use the more specific \tgtwos that are displayed on Play when viewing apps.

In our dataset, $\sim$65\% of reviews are non-English. Since our classifiers work on English texts, we leverage Google's Translation API
to translate reviews in 24 non-English languages to English (see~\autoref{sec:translated_languages}  for the full list)~\cite{bapna2022building}. The final set of English reviews (including translations) constitutes 98\% of the initial dataset and contains 1.9B reviews for 160K apps. These reviews come from all \googleplay categories and 445 \tgtwos, with representation from more than 200 countries and territories in the world.

Applying our \approach to this large review dataset, we identified 12.3M privacy-related reviews, which were organized into 227 themes with at least 5K reviews. To our knowledge, this constitutes the largest privacy-review dataset ever evaluated. In the following sections, we analyze these privacy reviews in depth across dimensions.

\subsection{Metrics}
\label{sec:privacy_metrics}
Hereafter, we rely on two core privacy metrics.
We first define the percentage of privacy reviews (\ppr): \[PPR = \frac{Number\ of\ privacy\ reviews * 100}{Total\ number\ of\ reviews}\]
When we compute \ppr over {\em all} 12.3M privacy reviews, we notate it as $PPR_{(all)}$. We compute $PPR_{(country)}$, where both the numerator and denominator are limited to one country. Similarly, we use  $PPR_{(app-type)}$ for the fraction of reviews within a particular app type that discuss privacy. 

\ppr{} is not applicable to \themes, only privacy reviews have \themes. To denote the ratio of a theme across all privacy reviews, we define percentage of theme privacy reviews (\ptpr): \[PTPR = \frac{Number\ of\ privacy\ reviews\ in\ a\ theme * 100}{Total\ number\ of\ privacy\ reviews}\]
\ptprc{(theme)} is often used to denote the theme of interest.

\subsection{Analysis Techniques}

We visualize data and 
report descriptive statistics to make 
observations. We use simple hypothesis tests, 
clustering, and regressions 
to examine trends in the 
underlying data. 

To understand if various subsets of privacy reviews 
significantly increase or decrease over time, we first perform a KPSS (Kwiatkowski–Phillips–Schmidt–Shin)
test to check if the \ppr/\ptpr time series (30 day intervals)
is stationary over the long term~\cite{kwiatkowski1992testing}.
We then fit a linear regression for 
each non-stationary item (e.g., \theme), 
to check if the slope estimate is statistically 
significant (i.e., \ppr/\ptpr changes over time). Items that 
do not meet these requirements are assumed to not consistently 
trend.
Further, we calculate the 
average two-year change in 
\ptpr/\ppr over the 10 years with a sliding 
window stride of 14 days.\footnote{We considered 
annual change but found two years 
to be more fluctuation resistant. Note that we do not
report the compounded rates.} Average two-year change 
mirrors regression results 
for every time-series analysis we 
conducted, and all items with significant KPSS values 
also had significant slope estimates.
Thus, for increased clarity, we only 
report average two-year change.

When elaborating on subsets of reviews (mostly obtained through 
the intersection of \themes, country, time, and \tgtwo),
we quantify the most prominent \issues and qualitatively assess a random 
sample of 
reviews to confirm. 
Our goal is to use examples to 
paint a more detailed picture of the 
\issues and \themes our \approach generates. 

\subsection{Ethical considerations} 
App reviews are public \omer{(and are denoted as such during submission}\begin{usenix}
\footnote{\url{https://web.archive.org/web/20230515141330/https://play.google.com/about/comment-posting-policy/?hl=en-US}}),
\end{usenix}
\begin{arxiv}
\cite{google2023commentpolicy}),
\end{arxiv}
and users submit reviews with the 
intention of sharing their views 
with other potential users. 
However, we take additional precautions before the review data is analyzed. 
First, all user identifiers (such as account IDs, emails, device information, etc.) are removed from 
the reviews dataset. 
Second, only reviews from apps with
at least 10K installs and at 
least 1K reviews are included. 
\omer{Third, inline with recent suggestions, 
we paraphrase all quotes
reported in this paper~\cite{kohno2023ethical}.}
Additionally, 
we do not disclose
app package names to prevent user 
deanonymization when joined with other sources.
Our dataset did not leave Google's premises, ensuring compliance with Google's terms of service. 

Our work uncovered reviews that contained ads for spying services. 
We disclosed our findings to Google Play and they removed 
these harmful reviews before publication.

\subsection{Limitations}

Our study is an observational one, 
i.e., we don't control who 
leaves privacy reviews when. 
The resulting selection bias is 
common in other work~\cite{mislove2011understanding, ruths2014social}; 
however, the bias we observe in this large-scale study of real-world 
data is likely different from the biases common in survey 
or interview studies, offering a different view of similar research questions. 

Each review, taken by itself, is specific to 
the application it is left on. This might mean 
our dataset is too specific and not generalizable. 
\nina{We argue that the sheer number of apps, and observed similarities within large groups of apps, helps to
smooth out this effect, and thus does reveal 
common themes among apps and countries.}
In isolation, feedback is 
specific; collectively, stories emerge. 
\sai{
We rely on Google Translate APIs to 
translate non-English reviews. This API has been thoroughly evaluated, even on low-resource (and low-volume) languages, and has F1 quality scores of $>$97\% for all languages with $>$2M native speakers (see Appendix E in~\cite{bapna2022building}).
Nonetheless, any translation 
errors might influence our results,
though we believe the impact is minimal since we focus on 25 widely spoken languages.
}

Our work focuses on Android users; however, other 
sizeable platforms exist, raising generalizability concerns depending 
on the ratio of android users in a country.
We note that Android is the most popular mobile operating system 
in the world and tends to be even more popular in non-WEIRD 
countries.\begin{arxiv}
\cite{androidios2023marketshare}.
\end{arxiv}
\begin{usenix}
\footnote{\url{https://worldpopulationreview.com/country-rankings/iphone-market-share-by-country}}
\end{usenix}
Coincidentally, these countries receive 
the least \sp research attention, 
some of which appear in our dataset but have 
received no prior privacy-focused academic interest~\cite{hasegawa2023survey}, 
making our work instrumental in addressing this gap.
We also note that, recent work has not found 
differences between Android and iOS users' privacy
sensitivities~\cite{abrokwa2021iosvandroid}, 
suggesting our results may provide hints 
beyond Android users. We leave the exploration 
of this hypothesis to future work.

\begin{figure}[t!]
    \centering
    \includegraphics[width=.4\textwidth]{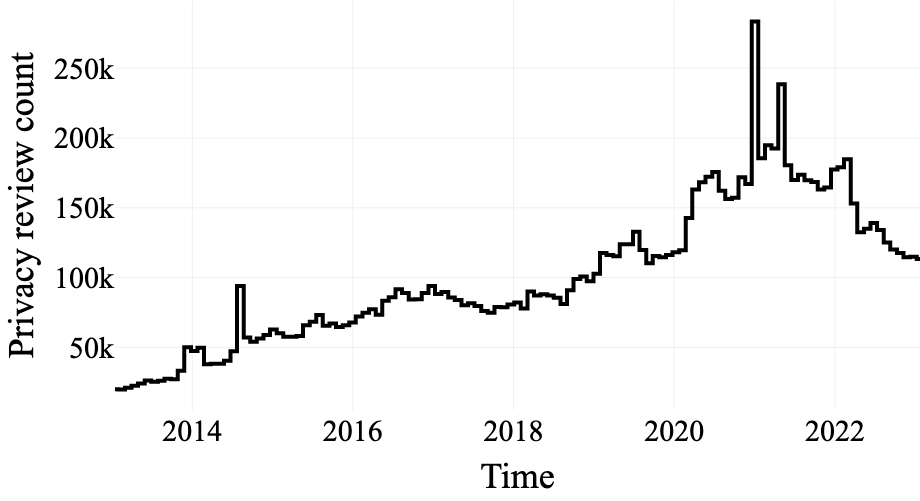}
    \caption{Privacy review counts per 30-day intervals.}
    \label{fig:number_of_privacy_reviews}
\end{figure} 

\section{Aggregate Growth}

\begin{figure}[t]
    \centering
    \includegraphics[width=.4\textwidth]{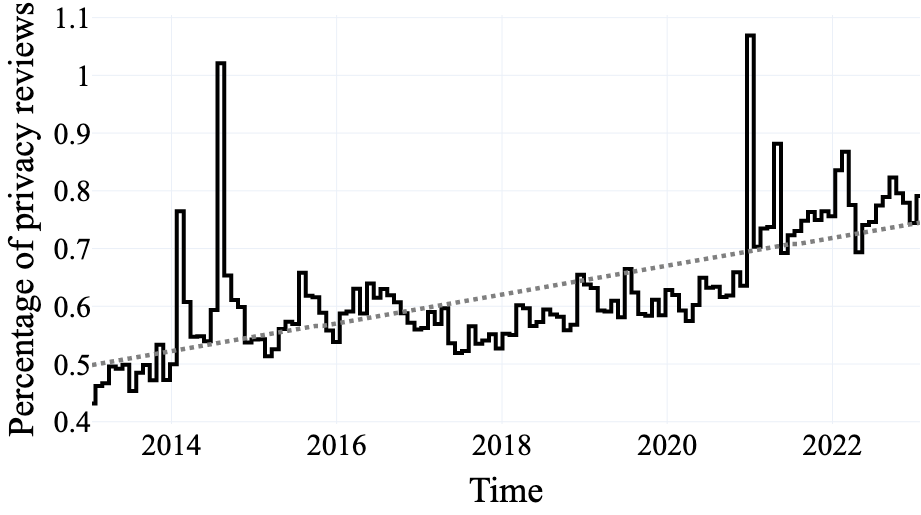}
    \caption{Percentage of privacy reviews (\pprall) per 30-day intervals. Dotted line is the fitted regression.}
    \label{fig:percentage_of_privacy_reviews}
\end{figure}

We start our dive into the 12.3M privacy reviews with an initial look at the aggregate data over the last decade. 
We first examine how \omer{privacy-related} 
reviews have grown over time. The absolute volume of privacy reviews, in 30-day intervals, is shown in \autoref{fig:number_of_privacy_reviews}. \autoref{fig:percentage_of_privacy_reviews} shows $PPR_{(all)}$ using the same intervals. Both figures trend upward over time, 
albeit non-uniformly, 
indicating that privacy reviews have increased over the last 10 years. Fitting a regression line on \pprall shows a significant increase
($R^2=0.44$, $p<0.001$, KPSS $p\leq0.05$). \pprall grows from 0.5\% 
to 0.8\% ($4.7\times$ increase in volume), a 9.3\% relative growth every 2 years on average. 
\sai{To contextualize this growth in privacy reviews, we also fit a regression line on the absolute volume of all (both privacy and not-privacy) reviews in 30-day intervals, and see that reviews have generally increased ($R^2=0.71$, $p<0.001$, KPSS $p\leq0.05$) during the same period from 6.5M to 24.5M per month ($3.7\times$ increase in volume). This shows privacy review volume is increasing faster than overall review volume.}

\begin{table}[t]
\footnotesize
\begin{tabular}{r l}
\toprule
\midrule
Date & Event \\
\midrule
February 2014 & Privacy reaction to the sale of a messaging app
\begin{arxiv}
\cite{facebook2014sec} \\
\end{arxiv}
\begin{usenix}
(\href{https://www.sec.gov/Archives/edgar/data/1326801/000132680114000010/form8k_2192014.htm}{link}) \\
\end{usenix}

February 2014 & Surge of a ``secure'' messaging app
\begin{arxiv}
\cite{telegram} \\
\end{arxiv}
\begin{usenix}
(\href{https://www.theverge.com/2014/2/25/5445864/telegram-messenger-hottest-app-in-the-world}{link}) \\ 
\end{usenix}

February 2014 & Proliferation of Talking Angela hoax
\begin{arxiv}
\cite{angela} \\
\end{arxiv}
\begin{usenix}
(\href{https://www.usatoday.com/story/news/nation-now/2014/02/20/talking-angela-app-scare-hoax/5635337/}{link}) \\ 
\end{usenix}

August 2014 & Large company forcing messenger download
\begin{arxiv}
 \cite{facebook2016forceddownload} \\
\end{arxiv}
\begin{usenix}
(\href{https://www.theguardian.com/technology/2016/jun/06/facebook-forcing-messenger-app-explainer}{link}) \\ 
\end{usenix}

Jan 2021 & Surge of another ``secure'' messaging app
\begin{arxiv}
\cite{signalmessaging} \\
\end{arxiv}
\begin{usenix}
(\href{https://www.forbes.com/sites/kateoflahertyuk/2021/01/11/how-to-use-signal-the-awesome-whatsapp-alternative/?sh=5fe8d91c324b}{link}) \\ 
\end{usenix}

Jan 2021 & Surge of privacy-focused search engine
\begin{arxiv}
\cite{duckduckgo} \\
\end{arxiv}
\begin{usenix}
(\href{https://www.usatoday.com/story/tech/2021/01/18/search-engine-duckduckgo-increases-traffic-google-competitor/4202556001/}{link}) \\ 
\end{usenix}

Jan \& May 2021 & Large messaging app privacy policy change
\begin{arxiv}
\cite{whatsapppolicychange} 
\end{arxiv}
\begin{usenix}
(\href{https://www.nytimes.com/2021/01/15/technology/whatsapp-privacy-changes-delayed.html}{link})
\end{usenix}
\\

\midrule
\bottomrule
\end{tabular}
\caption{Widely discussed privacy events}
\label{tab:event_table}
\end{table}

\begin{table*}[!ht]
    \footnotesize
    \centering
    \begin{tabular}{L{14.4em} c R{2.5em} L{24em} c}
        \toprule
        \midrule
        \textbf{Privacy Theme} & \textbf{Trend} & \textbf{Reviews} & \textbf{Short Summary} & \textbf{\# of Country Top-5 appear.} \\
        \midrule
        \texttt{Data Deletion} & $\Uparrow$ & 893K & Data deletion requests, data misuse, inability to delete data. & 47 \\
        \texttt{Privacy Concerns} & $\leftrightsquigarrow$ & 541K & Vague privacy protection and concerns. & 36 \\
        \texttt{Data Theft} & $\Uparrow$ & 450K & Data stealing, sharing, and leakage; unauthorized data access & 21 \\
        
        \texttt{Password Protection} & $\downarrow$ & 367K & Password protection for apps and user data. & 18 \\
        \texttt{Call Recording} & $\uparrow$ & 331K & Recording of phone call conversations. & 20 \\
        \texttt{Fingerprint Matters} & $\leftrightsquigarrow$ & 325K & Fingerprint scanners and their handling of the bio-metrics. & 17 \\
        \texttt{Excessive Permissions} & $\Downarrow$ & 322K & Asking for excessive \omer{privacy-sensitive} permissions. & 12 \\
        \texttt{Personal Information Privacy} & $\leftrightsquigarrow$ & 315K & Personal information access/usage. & 4 \\
        \texttt{Location Access Concerns} & $\leftrightsquigarrow$ & 313K & Location data collection and sharing. & 20 \\
        \texttt{Unneeded Camera Access} & $\downarrow$ & 311K & App accessing camera/microphone without permission. & 15 \\
        
        \texttt{Unneeded Access} & $\downarrow$ & 224K & Requesting \omer{privacy-sensitive} permissions (e.g., contacts). & 7\\
        
        \texttt{Content Hiding} & $\Downarrow$ & 223K & Efficacy discussion of content hiding features.  & 11 \\
        \texttt{Unauthorized Account Access} & $\uparrow$ & 220K & Unauthorized access to accounts, mobile devices, \ldots & - \\ 
        \texttt{Spying Concerns} & $\downarrow$ & 216K & Games mimicking users' considered spying. Surveillance. & 5 \\
        \texttt{App Locking} & $\downarrow$ & 205K & Privacy protections provided by app locking. & 3\\
        
        \texttt{Unwanted Data Collection} & $\leftrightsquigarrow$ & 200K & App collects data and sells them. & 2\\
        \texttt{Tracking Concerns} & $\leftrightsquigarrow$ & 190K & Tracking users phone number, location, and activity. & 2\\
        \texttt{Data Usage Concerns} & $\leftrightsquigarrow$ & 142K & Unwanted data usage patterns. & 3\\
        \texttt{Chats Privacy} & $\leftrightsquigarrow$ & 138K & Admiration for or the need to have private chat feature. & - \\
        \texttt{Information Privacy} & $\leftrightsquigarrow$ & 138K & App is not upfront about its functionality. & - \\
        \midrule
        \bottomrule
    \end{tabular}
    \caption{Top 20 \themes in privacy reviews over the last 10 years.
    $\Uparrow/\Downarrow$ indicates average \ptpr change $>1\%$ in two years. $\uparrow/\downarrow$ indicates significant change (KPSS test $p\leq0.05$).
    $\leftrightsquigarrow$ indicates no detectable significant change (KPSS test $p>0.05$).}
    \label{table:top20privacythemes}
\end{table*}

\autoref{fig:percentage_of_privacy_reviews} exhibits 
notable spikes in February 2014, August 2014, January 2021, and 
May 2021. \sai{Analyzing reviews during these periods, 
we identified a few apps contributing to these spikes. These apps had well-publicized 
events (\autoref{tab:event_table}) that heightened user privacy concerns or increased privacy awareness, 
resulting in the apps seeing a $2\times$ to $25\times$ increase in privacy review volume. Reviews of other \themes and \tgtwos stay 
relatively stable (see~\autoref{fig:items_over_time}).
}

\takeaways
Over the last decade, global privacy reviews have increased in absolute numbers and in \ppr, exhibiting a biannual relative growth rate in \ppr of 9\%.

\section{Trends in Privacy \Themes}

We now focus on our first research question (RQ1), asking which privacy themes are raised in reviews. \autoref{table:top20privacythemes} shows the top 20 themes by 
volume. We indicate whether the trend 
is generally increasing, decreasing or staying the same, 
and briefly summarize the review content. The last column states the number of countries where this theme appears among the top 5 themes, capturing how widespread an issue is globally.

We see that Data Deletion is the top theme worldwide, that it has been increasing over the last decade, and that it is a top issue in 47 countries. We note that the top 3 privacy themes constitute 16\% of all privacy reviews, the top 50 constitute 65\%, and the top 200 cover 85\%. From a volume perspective, this indicates that rather than a small number of dominant themes, we see a broad set of privacy topics raised across the Play store. However, from a geographic perspective, only 10 themes are a top issue in more than 10 countries (\autoref{table:top20privacythemes}). This indicates that some themes, even if quite voluminous, may arise in a limited set of countries.

To address our second research question (RQ2), on how privacy issues evolve over time, we first determine which themes are either decreasing or increasing in significant way.
In~\autoref{fig:theme_regression_coefficients} 
we plot the average 2-year change for themes among the top 20 that experience statistically significant change over the last decade (KPSS $p\leq0.05$); stationary (non-changing) themes are excluded.  The temporal evolution of five sample themes is shown in~\autoref{fig:themes_over_time}. In Sections~\autoref{sec:permissions}, ~\autoref{sec:increasing-themes}, and~\autoref{sec:loan_apps}, we continue to address RQ1 and RQ2 by focusing on the specific themes incurring long-term changes.

\subsection{\Themes Decreasing in Prevalence} 
\label{sec:permissions}

\begin{figure}[t]
    \centering
    \adjustbox{width=.515\textwidth, center}{
    \includegraphics[width=.50\textwidth]{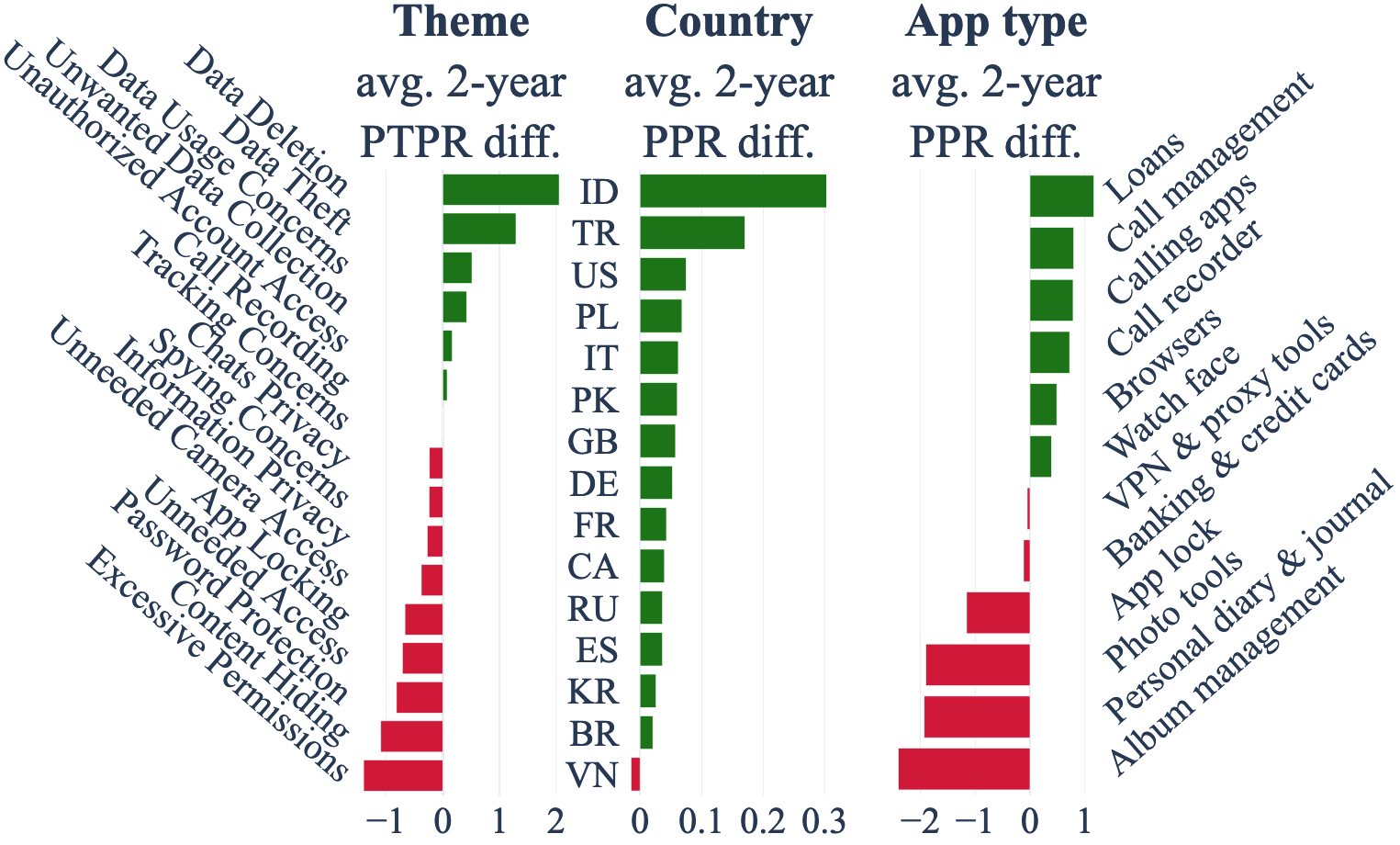}}
    \caption{Avg. 2-year \% point change 
    in theme (\ptpr), country (\ppr), and \tgtwo (\ppr). Green/red denotes increasing/decreasing trend (KPSS $p\leq0.05$).}
    \label{fig:yoy_change}
    \label{fig:theme_regression_coefficients}
    \label{fig:country_regression_coefficients}
    \label{fig:tg2_regression_coefficients}
\end{figure}

\paragraph{\omer{\Themes about privacy-sensitive permissions :}} Among the decreasing themes, (see~\autoref{fig:theme_regression_coefficients}, left-most column), 
we find several \themes related to \omer{privacy-sensitive} permissions, namely \texttt{Excessive Permissions, Unneeded Camera Access, Unneeded [Permission] Access}. We include \texttt{Location Access Concerns} in this batch, although it has been stable over time, as concerns about the location permission are a long-standing and much written about (see ~\autoref{sec:related-work}) privacy issue.

Our \approach captures nuance in reviewers' discussion of permissions, sorting \omer{privacy-sensitive} permissions reviews into distinct \themes. For instance, within the \texttt{Excessive Permissions} theme, reviewers complain about too many \omer{privacy-sensitive} permissions in general, 
without specifying which ones: e.g., 
``\textit{what's up with the permissions requests, 
why don't you listen to us and keep your hands off of our data}''.
In contrast, reviews labeled \texttt{Unneeded Camera Access} and \texttt{Location Access Concerns} exhibit concerns about camera and location permissions respectively. 
\texttt{Unneeded [Permission] Access} captures all lesser mentioned \omer{privacy-sensitive} permissions (47.5\% contacts, 9.3\% phone numbers, 7.5\% call logs, 
etc.).

\begin{figure}
    \centering
    \adjustbox{width=.51\textwidth,center}{
    \includegraphics[width=.50\textwidth]{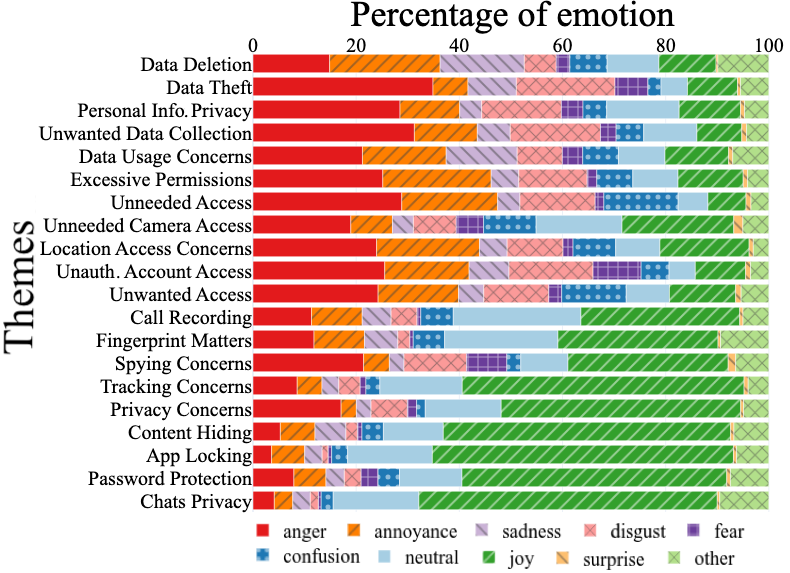}}
    \caption{Percentage of emotions in top 20 \themes, organized roughly by decreasing anger and increasing joy.}
    \label{fig:emotions_per_theme}
\end{figure}

We document here, for the first time, the dominant emotions associated with privacy reviews about app permissions. Unsurprisingly, we find that negative emotions (`Anger', `Annoyance', `Sadness', `Disgust', and `Fear'), account for 45--68\% of reviews most themes. 
(see~\autoref{fig:emotions_per_theme}). 
Qualitative analysis 
shows that `Anger' reviews commonly question the use case for \omer{privacy-sensitive} permissions 
and express disapproval, such as: 
``\textit{Horrible! it doesn't make sense to give 
permission for management of my email account, contacts, and images. All bad ideas.}''
Among our top 20 themes, the `Confusion' emotion appears most often with \omer{privacy-sensitive} permission \themes (7-14.4\%). The permission-related reviews associated with `Confusion' almost exclusively 
question \omer{privacy-sensitive} permissions but are less confrontational (e.g., 
``\textit{Why do you need identity and call information permissions?}'').

These \themes on \omer{privacy-sensitive permissions} have 
decreased significantly 
over time in \ptpr (see~\autoref{fig:yoy_change}): 
going from 19\% of privacy reviews to 8\%
between Feb 2013--Jan 2023. 
We hypothesize several explanations: 
(1) ongoing privacy enhancements in 
Android permissions (e.g., run-time permissions in version 6,
restricting background usage in version 10, 
one-time grants in version 11);
(2) Android's efforts in urging developers to reduce unnecessary permission requests~\cite{peddinti2019permissions}; and/or (3) the rise of other privacy concerns.

\begin{figure*}[t]
    \centering
    \includegraphics[width=.90\textwidth]{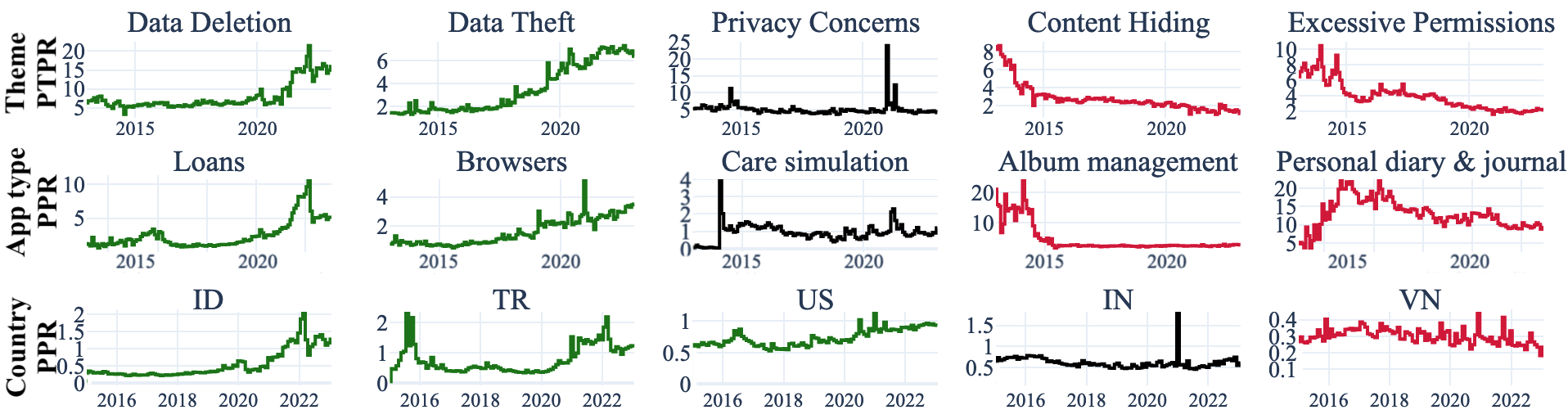}
    \caption{Selected \themes, \tgtwos, and countries over time. Green/red denotes increasing/decreasing trend (KPSS $p\leq0.05$)}
    \label{fig:items_over_time}
    \label{fig:themes_over_time}
    \label{fig:country_timeplots}
    \label{fig:tg2_timeplots}
\end{figure*}

\autoref{fig:themes_over_time} shows a brief 
increase in reviews \omer{about privacy-sensitive} permissions around 2016; 
we speculate this relates to increased 
visibility of permissions with the introduction of runtime 
permissions in October 2015.
The lag may be due to new Android versions being adopted over multiple years~\cite{Mahmoudi2018androidversion}.

\takeaways 
The fraction of privacy reviews related to \omer{privacy-sensitive} permissions has decreased from 19\% to 8\%, in the last 10 years.

\paragraph{Themes related to device sharing:} Other themes exhibiting significant decline in PTPR are \texttt{App Locking, Content Hiding} and \texttt{Password Protection}. These themes collectively have dropped from roughly 18\% of privacy reviews in Feb 2013 to 5\% as of Jan 2023.

We investigated the app types of these reviews, and found the reviews were predominantly for \textit{App lockers/hiders}, 
\textit{Album management}, and 
\textit{Personal diary \& journal}. Such apps typically allow protection of private content via password controls that are especially useful when a device is shared. 
South Asian countries contribute the most to 
these \themes (e.g., India makes up 29.4\% 
of \texttt{Content Hiding} reviews), and device sharing is 
a known challenge in these 
countries~\cite{sambasivan2018privacy}). 
We speculate increasing smartphone adoption---and 
thus reduced device sharing---might explain the decline among 
south Asian countries~\cite{india2022smartphonepenetration}. 
However, more research is needed to confirm. 
We further explore these \themes 
and \tgtwos in \autoref{sec:securityprivacy}. 

\subsection{\Themes Increasing in Prevalence} 
\label{sec:increasing-themes}
The top four themes that have shown statistically 
significant increases over time are:
\texttt{Data Deletion}, \texttt{Data Theft}, \texttt{Data Usage Concerns}, and \texttt{Unwanted Data Collection}. These \themes have risen in aggregate
from 10\% of privacy reviews in early Feb 2013 to 27.3\% by Jan 2023. Collectively, these themes focus on PII and user-generated personal data. 
Our analysis methods discern the nuanced difference between PII and \omer{privacy-sensitive} permission discussions. 

Themes in this group capture feedback in nuanced ways. 
\texttt{Data Deletion}-\themed reviews are almost exclusively requests to delete personal data (discussed in~\autoref{sec:loan_apps}). 
In contrast, \texttt{Data Theft}-\themed reviews express concern about data being stolen or an account hacked.
\texttt{Data Usage Concerns} primarily refers to sharing data with third parties, and
\texttt{Unwanted Data Collection} reviews express concerns with the app requesting unnecessary data (e.g., for showing ads).

These reviews occur in two styles. Many reference precise data types: national IDs (e.g., SSN, TCKN), financial data (e.g., credit cards), personally identifiable information (e.g., emails), or user-generated content (e.g.,
``\textit{Requires EVERYTHING about you including your social security number to scan lottery tickets. Why wouldn't you
use your local store instead of exposing yourself to a risk 
of data breaches?}'').
Other reviews in these themes discuss personal data vaguely, without reference to specific personal data types (e.g., 
``\textit{Warning! The app collects a lot of data. Personal data, device and usage-related data\ldots The app then sells it}'').

Understandably, the most common emotion with these 
reviews is `Anger' (15-35\% of privacy reviews per theme). 
\texttt{Data Theft} also contains a larger 
rate of `Fear' (6.4\% of privacy reviews). 
Interestingly, these reviews are not strictly negative; 
9-12\% of privacy reviews on these personal data \themes exhibit `Joy.' 
Some express gratitude for apps that protect data, such as 
``\textit{This app locks your personal stuff. Great app.}''

\takeaways
Over time, an increasing share of privacy reviews focus on personal information (e.g., PII). 
Our analysis uncovers privacy discussions about different types of personal data, such as \omer{privacy-sensitive} permissions vs PII.
\subsection{The rise in Data Deletion and Data Theft}
\label{sec:loan_apps}

Since \texttt{Data Deletion} and \texttt{Data Theft} are our two largest themes, and both show a statistically significant increase in volume over time, we now look at what the contributing factors may be.  Our analysis enables us to break down our data along multiple dimensions, thus we looked at per-country contributions for these two topics and found that Indonesia (disproportionately) contributes 40\% of the reviews in these two categories, whereas all other countries contribute less than 8\% each. We also saw that Indonesia has the fastest \ppr growth of any country, with 76\% relative biannual growth and the third-most privacy reviews of all countries by volume. Considering only the  \texttt{Data Deletion} and \texttt{Data Theft} themes for Indonesia, we next considered \tgtwos. We uncovered that 44\% of these reviews came from financial loan apps, with other \tgtwos contributing less than 0.8\% each.

Many \texttt{Data Theft} reviews complain that the app collects PII (social media accounts, photos of IDs), and then either loans are denied without reason, or threats are made leveraging their PII. One reviewer writes
``\textit{The app stole my data, including my facebook account. 
It told me that the loan process would be easier if I input all of my data.
My loan was still denied even though I have a good payment history. 
Watch out people.}''
Another reviewer says 
``\textit{[COMPANY] pretends you'll get a loan with no hiccups.
They say they just need your photo identity. 
If you're even a day late, their debt collector will harass you and 
threaten to leak your data (photo identity) to social media.}''

Most \textit{Data Deletion} reviews  
we manually examined ask for personal information to be deleted because the user's loan was rejected. Some claim payment was requested when a loan was never applied for or received. For instance, one reviewer from Indonesia exclaims,
``\textit{Didn't receive a penny! Why do you threaten me with calling all of my contacts if I don't pay? DELETE ALL OF MY DATA PLEASE.}''

Munyendo et al.\ conducted a user study with 20 participants in 
Kenya and reported that loan apps were calling 
the contacts of people who had not applied for loans~\cite{munyendo2022kenya}. Our NLP analysis pipeline was able to uncover that this issue, first noted in Kenya, is in fact a more widespread problem. In addition to Indonesia, we observe many \texttt{Data Deletion} 
and \texttt{Data Theft} reviews on loan apps from India,
Mexico 
and Thailand. In our data, 
Kenya contributes only a small portion of these reviews, 
which may reflect population size and/or selection 
bias in who leaves reviews on Google Play. 

\autoref{fig:country_regression_coefficients} indicates that the loans \tgtwo is a growing concern, with the highest growth in privacy reviews among \tgtwos. 
In 2022, many such abusive financial loan apps 
were removed from the market\begin{usenix}
.\footnote{\url{https://techcrunch.com/2022/11/18/google-clamps-down-on-illegal-loan-apps-in-kenya-nigeria/}}
\end{usenix}
\begin{arxiv}
\cite{annie2022loans}.
\end{arxiv}
Subsequently in 2023 Google announced a policy update prohibiting personal loan apps from accessing sensitive data such as photos and contacts\begin{usenix}
.\footnote{\url{https://support.google.com/googleplay/android-developer/answer/9876821}}
\end{usenix}
\begin{arxiv}
\cite{playloanappspolicychange}. 
\end{arxiv}
Since our data collection ended in early 2023, we are unable to measure the impact of this policy.

\takeaways
We showed how our ability to break the data 
down along multiple dimensions 
(country, time, theme, app type) 
can be useful in understanding trends. 
This evaluation also illustrates how text analysis 
and user studies can complement 
each other: when a prior user study~\cite{munyendo2022kenya} 
identifies an unexpected issue, it can be 
followed up with text analysis to 
quickly examine the issue's geographic spread.

\section{Trends in Countries}
\label{sec:country-trends}

We now focus on our third research question (RQ3), examining how privacy themes discussed in reviews vary around the globe. Our 12.3M review data set includes more than 200 countries or regions. The top 10 contributing countries are
the U.S. (15.4\%), 
India (12.1\%),
Indonesia (10.5\%), 
Brazil (7.7\%), 
\turkey (4.5\%), 
Russia (4.4\%),
Mexico (4.2\%), 
Germany (3.0\%), 
Pakistan (2.6\%), and
United Kingdom (2.4\%).  
These 10 countries together contribute 66\% of privacy reviews, with the top 33 countries supplying 90\%. Interestingly, 1) only two European countries are among the top 10 contributors by volume, 2) the top 33 countries come from every continent, and 3) there is a long-tail of privacy review percentages coming from more than 150 countries. Overall, while people from all corners of the world write about privacy, we find that 33 countries dominate this discussion.
\sai{This distribution of privacy reviews across countries largely follows the trends in all (both privacy and not-privacy) reviews submitted, except for few minor shifts in country rankings. For instance, Brazil is second in terms of review volume, but drops to fourth with  privacy reviews. Similarly, South Korea drops from eighth to 13th, while Spain jumps from 15th to 11th. Nonetheless, the top countries that contribute the highest number of reviews are also the ones that contribute the most privacy reviews.}

In the rest of this section, we first cluster countries (\autoref{sec:clustering}) to understand where broad commonalities do and don't exist. We then take a closer look at countries with outlier patterns (distributions of privacy topics discussed) in~\autoref{sec:unique_countries}, and other anomalous patterns such as high growth rates in \turkey (\autoref{sec:turkey}) and unusual reviews about spying in Nigeria (\autoref{sec:nigeria}). It is beyond the scope of this paper to carry out further detailed world-wide country comparisons.

\begin{figure}[t!]
    \centering
    \includegraphics[width=.52\textwidth]{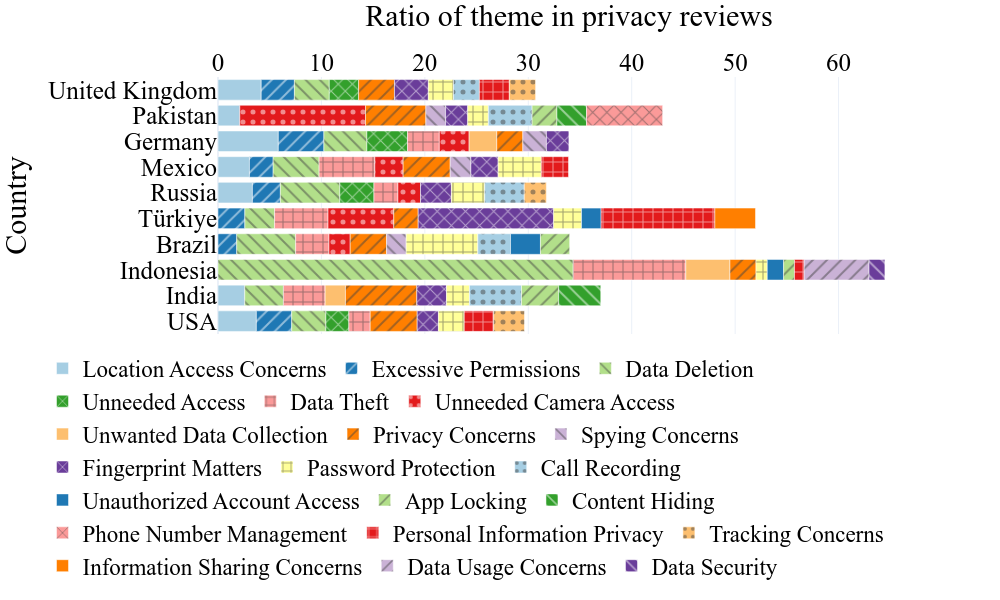}
     \caption{Top 10 themes per country (PTPR), showing variation in theme popularity across countries.}
    
    \label{fig:top_10_themes_per_county}
\end{figure}

\subsection{Clustering Countries}
\label{sec:clustering}

\nina{Earlier research \cite{cho2018collective,redmiles2019should,herbert2022world,anaraky2021privacyculture}} \nina{has hypothesized and examined whether the influence of culture and geographic proximity changes 
privacy attitudes}. While loose correlations are sometimes observed (e.g.,~\cite{redmiles2019should,herbert2022world}), often it is difficult to identify strong predictors because there are so many factors that influence privacy~\cite{cho2018collective,anaraky2021privacyculture}. We \nina{explore} the potential connection between geographic proximity and privacy opinions from a different angle: we examine expressed opinions, from people living in a country, over a range of privacy topics. We do not study underlying factors, such as culture; instead we offer direct data summaries of user-provided texts.

\begin{figure}[t!]
    \centering
    \includegraphics[width=.48\textwidth]{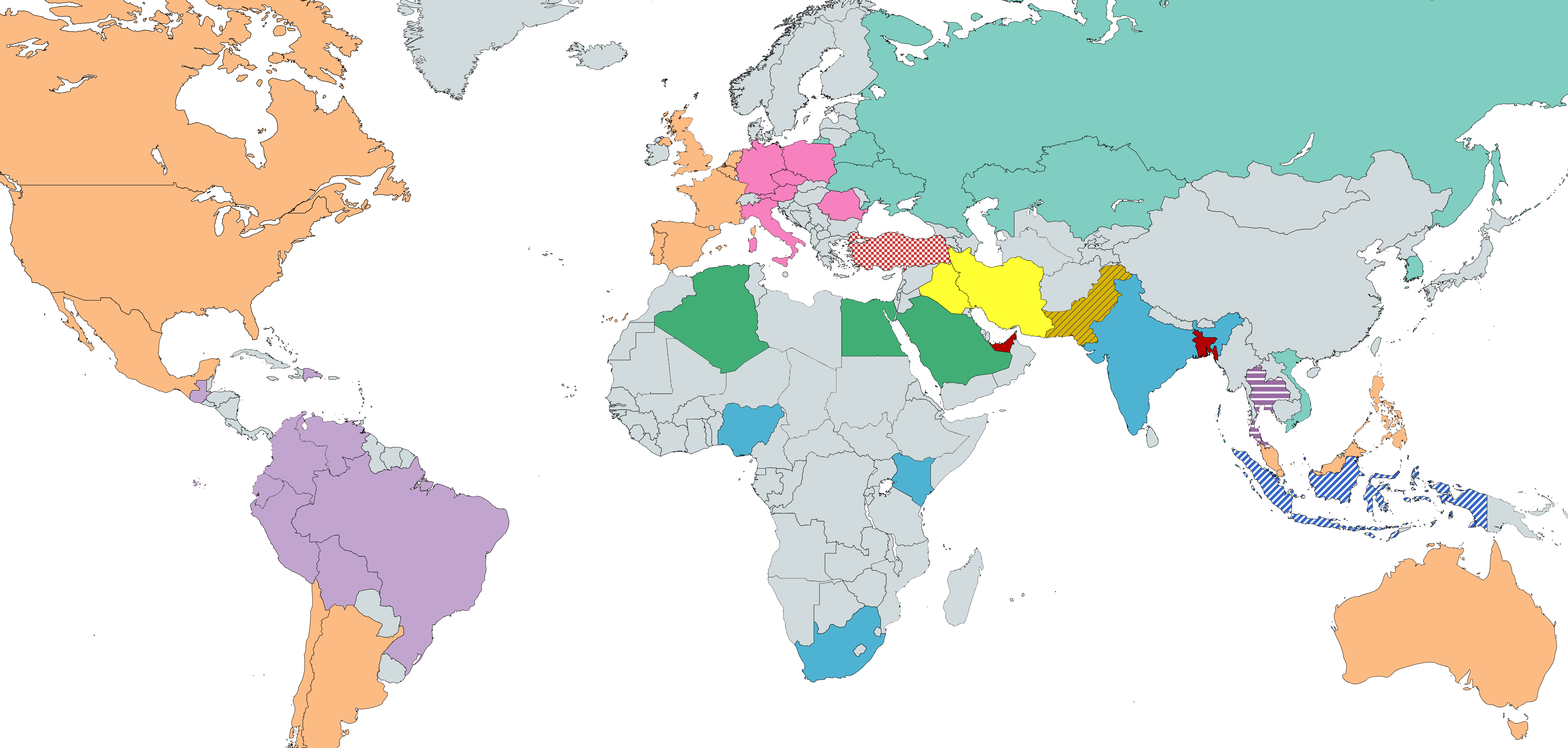}
     \caption{Countries clustered according to distribution of privacy themes discussed in app reviews.}
    
    \label{fig:world_map}
\end{figure}

\nina{We investigate the question of whether countries that are geographically close discuss similar app privacy issues, in an exploratory fashion}. We cluster countries based on prevalence of privacy \themes discussed.\footnote{While this is 
limited by those who chose to 
write reviews, it nevertheless is 
based on sizeable inputs from each 
country.} As intuition for this approach, we provide a visualization in \autoref{fig:top_10_themes_per_county} of the distribution of themes for the top 10 countries contributing the most privacy reviews. (We only include top 10 themes per country to facilitate readability.) 
We see that the U.S. and U.K. have quite similar distributions, while Indonesia, \turkey and Pakistan have distinct patterns.

We now use these topic distributions to cluster countries. Specifically, we represent each country as a vector of size 50, where each element is the \ptpr for a (country, \theme) tuple, for the top 50 themes across the entire dataset.  
We then apply hierarchical clustering with complete linkage~\cite{murtagh2012algorithms}. 
The resulting clusters are shown on a world map in~\autoref{fig:world_map}.
A detailed breakdown can be seen 
in\begin{usenix}
~\omer{Figure 12 of the extended paper~\cite{akgul2024reviews}.}
\end{usenix}
\begin{arxiv}
~\omer{\autoref{fig:cluster} in~\autoref{app:cluster}.}
\end{arxiv}

We see from this map that the question of whether geographically close countries write about the same privacy issues yields mixed results. On one hand, Russia, Ukraine, Belarus and Kazakhstan all cluster together (teal). We also see that most of the Anglo countries, namely the U.S., Canada, Australia and the U.K., are in the same (light orange) cluster. However, this light orange cluster also contains Mexico, Argentina, Spain, Portugal, and France. Interestingly, European countries divide between two clusters, with eastern European countries, Germany, and Italy in their own cluster (pink). Similarly, the Middle East and North Africa splits across three main clusters: \turkey is alone, Iran and Iraq cluster together (yellow), and Saudi Arabia, Algeria, and Egypt cluster together (green). A detailed examination of our dendrogram shows that the Iran-Iraq cluster is surprisingly far away from the Saudi Arabia-Algeria-Egypt cluster. Asian countries exhibit wide diversity. India clusters with some African countries (South Africa, Kenya, and Nigeria), and not with Pakistan or Bangladesh. Interestingly, we note that Indonesia, Pakistan, Thailand and \turkey appear unique (cluster size 1).

\takeaways
Geographical proximity does not reliably indicate whether countries discuss the same set of privacy topics.

\subsection{Unique distributions of privacy themes}
\label{sec:unique_countries}

As noted above, our clustering identified four countries that are alone in clusters of size 1: Indonesia, Pakistan, Thailand, and \turkey. We also found two small clusters of only two countries: Bangladesh with the United Arab Emirates, and Iran with Iraq. In~\autoref{tab:top-issue-unique-countries}, we list the most frequently discussed issues for these countries. 

\begin{table}[]
    \centering
    \footnotesize
    \begin{tabular}{l l} 
    \toprule
    \midrule
      Country   &  Most Discussed Privacy \Theme \\ \midrule
      Pakistan  &  \texttt{Unneeded Camera Access} \\
     Indonesia  &  \texttt{Data Deletion} (mostly in loan apps) \\
     \turkey & \texttt{Fingerprint Matters} \\
     Thailand & \texttt{Data Deletion} (broadly across app types)\\
      Bangladesh \& UAE & \texttt{VPN Matters} \\
      Iran \& Iraq & \texttt{Data Deletion} (comms \& social media apps)\\
      \midrule
      \bottomrule
    \end{tabular}
    \caption{Countries with unusual \theme distributions}
    \label{tab:top-issue-unique-countries}
\end{table}

Pakistan is the country with the highest rate of \texttt{Unneeded Camera Access} (12.2\% of reviews from Pakistan). These reviews distinguish themselves from other countries by praising the abilities of \texttt{Hidden Camera Detection} (issue) apps, which claim to detect hidden cameras using device magnetometers (e.g., 
``\textit{the app detects all hidden cameras in your vicinity, I love this app}'').
Pakistan's focus on this issue could be a reflection of a higher adoption of this type of app. 

The reviews from Indonesia were discussed in \autoref{sec:loan_apps}. The reviews from \turkey are discussed in more depth in \autoref{sec:turkey}. Distinctively, Thailand exhibits a combination of the unique behaviors observed in reviews from Indonesia and \turkey.  

Bangladesh and UAE have a higher fraction of  
reviews categorized as \texttt{VPN Matters} than any other country. These reviews are overwhelmingly positive but predominantly non-specific, e.g.,
``\textit{Love this app. It is a great vpn.}''
We observe that 55.0\% of \texttt{VPN Matters} 
reviews are for apps that contain 
`Free' in the title. Prior work has 
found that ``free'' VPNs are
often misconfigured or outright 
malicious~\cite{ramesh2022vpnconsumer, perino2019long, ikram2016analysis}. 
A possible explanation for this \theme's relative 
prevalence could be the common use of VPNs to 
circumvent the relatively high rate of 
censorship in Bangladesh and UAE~\cite{2022internetfreedom}. 

For both Iran and Iraq the top \theme is \texttt{Data Deletion}, with the vast majority of reviews applying to communication and social media apps
(see~\autoref{sec:apps} for more on apps).

\subsection{\turkey}
\label{sec:turkey}

\turkey stands out as anomalous based on two metrics. First, its distribution of privacy topics does not cluster with any other country (cluster size 1). Second, \turkey exhibits anomalous PPR growth over ten years.  The middle 
column of~\autoref{fig:country_regression_coefficients} shows 
the average 2-year change for countries, among the 
top 20, with statistically significant growth or 
decline in PPR (KPSS $p\leq0.05$); stationary (non-changing) countries 
are excluded. \turkey shows an unusually large increase, with an
average 2-year relative growth of 62.1\% between Feb. 2015 and Jan. 2023 (\autoref{fig:country_timeplots} illustrates this).
\turkey also provides the 6th-most privacy reviews (502K) of all countries.

\autoref{fig:top_10_themes_per_county} shows
that \turkey has an unusual distribution of privacy 
themes;  \texttt{Fingerprint Matters}, \texttt{Personal Information Privacy} 
and \texttt{Unneeded Camera Access} appear more frequently than in other 
countries. 
The privacy reviews highlight user concerns around fingerprint 
collection (e.g., 
``\textit{My fingerprint was scanned on [date], 
the company has the responsibility of my fingerprint.}'')
and abuse (e.g., 
``\textit{If my fingerprint is used in something illegal, 
the app is responsible.}'').
Although biometrics such as fingerprints are not shared with apps 
directly\begin{usenix}
,\footnote{\url{https://source.android.com/docs/security/features/authentication/fingerprint-hal}}
\end{usenix}
\begin{arxiv}
\cite{biometrics2022android},
\end{arxiv}
reviewers express 
concern about the potential for such sharing, e.g., when a 
banking app verifies authorization via a fingerprint. 

In addition, we observed that 29\% of privacy reviews from \turkey include 
text we refer to as a ``disclaimer'': quasi-legal language asserting rights 
or claims. These disclaimers assert app developers' responsibility to safeguard information (e.g., 
``\textit{This app is nice but you responsible for any inappropriate use of my personal data.}'')
or longer texts that threaten legal recourse for violations:

\begin{quotation}
``\textit{This application was downloaded at [time] on [date]. I [give] no permission to share things (My photo,
T.C. ID number, password, etc.) with third parties. And if such thing happens, \dots the app bears sole responsibility and I will take legal action against them.}''
\end{quotation}

These disclaimer reviews appear for multiple \tgtwos, 
including call management/recorders, 
physical activity trackers, 
investment/cryptocurrency apps, 
VPNs, and even antivirus apps; 
many of these typically require access to 
sensitive permissions or resources such as 
app usage patterns, files, and network control. 
The disclaimer reviews hint at nuanced mental 
models in which reviewers worry that an app 
might violate the privacy it claims to protect.

Overall, we hypothesize that reviewers in \turkey use these disclaimers 
when they are uncomfortable with an app's privacy risks but feel compelled 
to use it, such that the disclaimer seems like the only protective option. 
Anecdotally, conversations with Turkish nationals suggest similar disclaimers 
circulate on messaging and social media apps with privacy 
protection promises.\footnote{Similar disclaimers have circulated elsewhere for 
years~\cite{facebook-hoax-2015,facebook-scam-2019}.
}
These disclaimers raise additional 
research questions: Why are these reviews 
seen so frequently from \turkey? Are reviewers 
who post these disclaimers unusually privacy-sensitive?
Do they believe the disclaimers 
have legal value? Do they employ other privacy protections?

\takeaways
Over 150K reviews in \turkey post 
quasi-legal disclaimers asserting privacy rights. We hypothesize a large-scale 
misunderstanding might cause such reviews.

\subsection{Nigeria}
\label{sec:nigeria}
Nigeria is the only country with \texttt{Spying Concerns} 
as the top theme. Surprisingly, many of these reviews are associated with `Joy,' and unlike most \texttt{Spying 
Concerns} reviews from other countries,
they aren't about 
games (see~\autoref{sec:spyingtom}). Manual inspection reveals 
that reviews for tracking related apps sometimes contain ads for spying services, for example:

\begin{quotation}
``\textit{I can't find enough words to 
thank [email]! I tried lots of 
times to spy on my 
spouse \ldots
to no avail. 
This guy is magic, within three hours, 
he gave me access the calls and messages 
of my spouse.}''
\end{quotation}

\Issues associated with \texttt{Spying Concerns}  
include \texttt{Spying on Spouse}, 
\texttt{Spying on Partner}, 
\texttt{Allows Unauthorized Calls Access}, 
and \texttt{Allows Unauthorized Messages Access}, 
suggesting these ads may primarily be targeted to intimate partner abuse perpetrators. 
Notably, many of these review 
ads use similarly formatted 
email addresses. 
Although we identified this pattern in privacy reviews, a simple regular expression\footnote{Omitted here due to potential harm; contact the authors for information.} 
across all reviews yielded more than 10K matches. 
A random sample of 100 matches yielded only three false positives.
Matching reviews appear on apps with 
\dev-specified
categories such as Lifestyle, Tools, 
Books, Communication, and Dating. 
Most are from Nigeria.
We disclosed this issue to Google Play, and these reviews were taken down.

A simple web search for some of the email addresses 
from these reviews reveals similar ads posted as comments/reviews 
across the web as well as posts about the (in)effectiveness 
of the advertised spying\begin{usenix}
,\footnote{\url{https://www.scamwatcher.com/scam/view/272748}}
\end{usenix}
\begin{arxiv}
\cite{spywarehacker},
\end{arxiv}
suggesting these services are not unique to Google Play, 
and may be of interest to 
the digital intimate partner violence research 
community~\cite{freed2018ipa, havron2019ipa, bellini2023ipa}.

\takeaways
We detected a subset of reviews offering spyware services, primarily from Nigeria. These were reported to Google Play and have been removed before publication.

\section{Stark Differences Across \TgTwos}
\label{sec:apps}

There are 445 \tgtwos in our dataset (functional categories displayed on Play app page), with
some invoking strong emotional reactions from users.
We address RQ4 by focusing on \tgtwos among the top 50 (by privacy review volume) that:
(1) receive a high rate of negative-emotion privacy reviews, or (2) receive overwhelmingly positive reviews.
We draw connections with previous research when possible (addressing RQ5), while also identifying areas that have been under-explored in the literature.

\begin{figure}[t]
    \includegraphics[width=.48\textwidth]{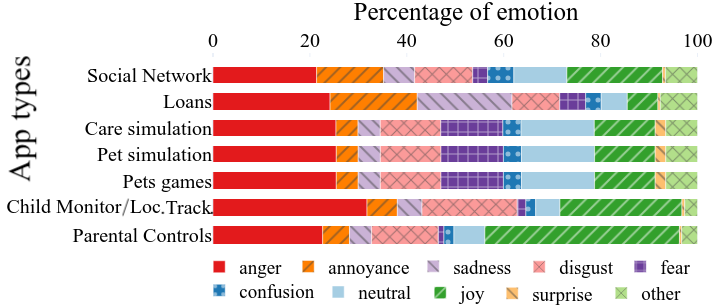}
    \caption{App types with highly negative privacy emotions. }
    \label{fig:emotions_per_tg2}
\end{figure}

\subsection{App types with negative emotions}

We define \tgtwos with strongly dominant negative emotions as those where more than 50\% of privacy reviews were assigned to any of `Anger', `Annoyance', `Sadness', `Disgust', or `Fear'
(see ~\autoref{fig:emotions_per_tg2})
Of the 25 \tgtwos that satisfy this condition, we highlight only a few below due to limited space (loan apps were already discussed in~\autoref{sec:loan_apps}).

\paragraph{Social media:}
These apps receive the second-most 
privacy reviews (775K) of all \tgtwos, and $\sim$57\% of these reviews are associated with negative emotions. 
Reviewers most 
frequently bring up \texttt{Invasion of Privacy}  
and \texttt{Data Deletion}. Non-specific comments, such as in  
\texttt{Invasion of Privacy}, include: 
``\textit{You invade the privacy of people. Our privacy is sold for profits. You let fake news be posted by foreign countries.}''
In \texttt{Data Deletion} reviews, users complain that they are unable to delete accounts,
posted content, 
or the app itself (some social media apps come pre-installed and are not removable~\cite{gamba2020analysis}). 
This latter feedback is more actionable for developers; for example, improving the UI to make data deletion controls more discoverable could help. The strongly negative emotions indicate users who are very frustrated with privacy properties of social apps.

\paragraph{Pet simulators:}
\label{sec:spyingtom}
Several popular `simulation' 
games that mimic users' facial and vocal expressions are 
grouped under the game categories 
\textit{Care Simulation}, \textit{Pet Simulation}, and \textit{Pets}. 
These \tgtwos have a total of 199K privacy reviews, and 60\% of these are associated with strong negative emotions.
Reviews left on these games are often assigned \themes including 
\texttt{Unneeded Camera Access}, 
\texttt{Spying Concerns}, 
and \texttt{Unauthorized Surveillance}. 
These reviews largely express concern that the `pet' is watching the user
through cameras embedded in its eyes. For example: 
``\textit{I don't recommend the app. She
is immensely dangerous as she has cameras for 
eyes. She captured my picture. I was playing with 
it at 3am and she said she would come to my 
house.}''
Though these reviews are based on 
misconceptions\begin{usenix}
,\footnote{\url{https://www.usatoday.com/story/news/nation-now/2014/02/20/talking-angela-app-scare-hoax/5635337/}}
\end{usenix}
\begin{arxiv}
\cite{angela}, 
\end{arxiv}
as these games 
are primarily intended for children, 
they often include 
heightened emotions. 
Reviews such as these were
prevalent in the United States, Brazil, India, Mexico, and Italy. They first appear in early 2014 and have not diminished since.

\begin{figure}[t!]
    \centering
    \adjustbox{width=.51\textwidth, center}{
    \includegraphics[width=.50\textwidth]{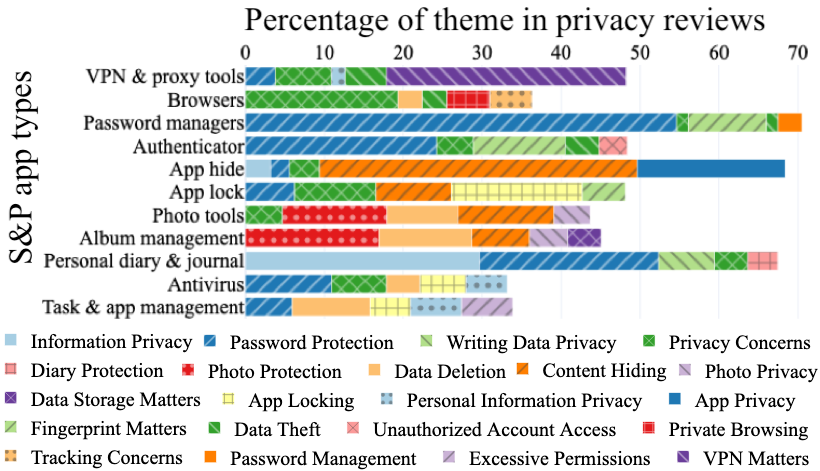}}
    \caption{Top \sp \tgtwos \themes (\ptpr).}
    \label{fig:sp_category_issues}
\end{figure}

\paragraph{Child monitors and location tracking:}
\textit{Child Monitors \& Location Tracking} and \textit{Parental Controls} apps also receive a lot of 
negative feedback, with 52\% of 135K privacy reviews associated with negative emotions. 
Generic \texttt{Privacy Concerns} 
is the top \theme; 
somewhat less common, but more specific, \themes 
include \texttt{Children Privacy Concerns} 
and \texttt{Monitoring Children}. 
Reviewers of these apps typically complain about being 
tracked by the app or object to the 
existence of this \tgtwo in 
general.\footnote{Because these reviews may 
have been by children, we provide 
no quotes.}
However, positive privacy reviews of 
child-monitoring apps are also 
common (36\% contain `Joy').
Positive reviews frequently praise 
the ability to \texttt{Track} 
and \texttt{Access Location}. 
Conflicts between children's privacy and 
parental supervision have been studied in a variety of 
contexts (e.g.,~\cite{cranor2014parents,wei2022anti}), including a 
study of 736 reviews of child-monitoring apps~\cite{ghosh2018safety}.
In that study, which explicitly aimed to study reviews by children, 
most reviews were negative.

Other work has reported on the use of 
these apps for intimate partner 
abuse~\cite{chatterjee2018spyware}. 
Our manual analysis did not identify any reviews 
explicitly acknowledging tracking adults, and we observe 
few reviews in related categories like \texttt{Spying concerns} or \texttt{Unauthorized surveillance}. This identifies a limitation of using our approach to complement user studies or surveys, namely that we cannot guarantee that reviews on a particular topic will be part of our data collection.

\subsection{Privacy Protective Apps: Quite Positive!}
\label{sec:securityprivacy}
One might intuit that when users write about privacy, 
they primarily complain, and hence those reviews are 
associated with negative emotions. Interestingly, 32\% 
of all privacy reviews are associated with `Joy' emotion.
Often these reviews express appreciation for a particular privacy feature (e.g., private chat). To explore this further, we identified \tgtwos in the top 50 with at least 25\% of reviews associated with `Joy' (i.e., privacy positive reviews). Of the 23 \tgtwos satisfying this condition, the majority were privacy- and security-focused. 

\begin{figure}[t]
    \centering
    \adjustbox{width=.49\textwidth, center}{
    \includegraphics[width=.50\textwidth]{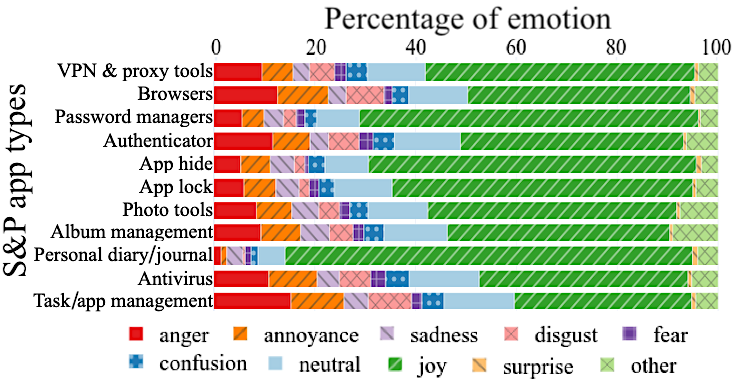}}
    \caption{Top security \& privacy app types with highly positive emotions.}
    \label{fig:sp_category_emotions}
\end{figure}

Below we summarize reviews from these privacy- and security-focused \tgtwos. Their dominant privacy themes are shown in~\autoref{fig:sp_category_issues}, and
\autoref{fig:sp_category_emotions} shows the emotion distribution.
At a high level, users appreciate the (perceived) protections
they receive from these apps. Our results confirm and extend prior work while also highlighting research gaps.

\paragraph{VPN \& proxy tools:}
These apps received 128K privacy reviews, and 53\% of these express 'Joy.' Reviewers broadly discussed \texttt{VPN Matters}, 
\texttt{Privacy Concerns}, 
and \texttt{Data Theft} 
\themes.  
\texttt{VPN Matters} reviews frequently consist of short, vague endorsements
(e.g., 
``\textit{Great VPN, among the best out there}'').
We looked at the fine grained issues within the VPN matters theme, such as, 
(\texttt{Best VPN}) and found the issue names similarly undescriptive.
Broad affection for VPNs has been reported in prior measurements~\cite{akgul2022vpn, dutkowska2022vpn}.
Within \texttt{Privacy Concerns}, we also see 
fairly generic positive reviews, grouped under issues like \texttt{Privacy Protection} 
(e.g., ``\textit{Love the app. I get decent speeds and excellent privacy.}'')
and \texttt{IP Protection}. 
These comments are similar to terminology appearing in 
influencer VPN ads~\cite{akgul2022vpn}.

Differently from some prior work, reviewers in our 
dataset do not commonly 
mention benefits such as protection against internet 
surveillance~\cite{namara2020emotional, ramesh2022vpnconsumer}, 
utility for censorship evasion~\cite{dutkowska2022vpn}, or lack of server-side logging~\cite{ramesh2022vpnconsumer}.
Further, while findings from prior work on VPNs tend to emphasize specific adversaries~\cite{akgul2022vpn, ramesh2022vpnconsumer, dutkowska2022vpn}, we don't observe \themes 
or \issues with this emphasis. As expected from prior work~\cite{story2021awareness}, reviews commonly show confused mental models, e.g., when one user claims a VPN  
``\textit{Hides my IP address from my internet ISP.}''

\paragraph{Browsers:}
These apps receive the fifth-most privacy reviews (332K) of all app types, 44\% of which express `Joy.' The majority of these privacy-positive reviews 
apply to 
smaller browsers that advertise privacy-focused design. 
Reviews praise \texttt{Private Browsing} modes or private browsers 
and built-in tracking protection (\texttt{Tracking Concerns}), 
e.g., 
``\textit{Amazing browser for your privacy, blocking trackers. They do not keep search records is just what people need recently}.''
Overall, users seem to appreciate browsers with enhanced privacy protections. 
Browsers are the only \sp \tgtwo with significantly increasing (KPSS $p\leq0.05$) privacy reviews: a two-year average growth of 0.5\% points in \pprc{Browsers}(46.5\% average relative growth).

\paragraph{Password Managers:}
Out of 27K privacy reviews, reviewers most frequently 
praise password managers' main function, 
\texttt{Password protection},
as well as using biometrics to 
unlock password storage (\texttt{Fingerprint Matters}).
Overall, \textit{Password Managers} are more 
frequently labeled with  `Joy' (67.2\%, compared to 5.6\% of `Anger'). 
Accordingly, we see nearly no complaints 
in top \themes. A minority 
complain that fingerprint authentication doesn't work,
and a smaller minority say they are unable to change their master password.

These complaints don't 
necessarily align with 
problems identified in prior work~\cite{pearman2019pm, ray2021pm, mayer2022pm}, and we observe few complaints about usability 
or interaction problems~\cite{huaman2021pm}. This 
discrepancy may be because reviewers likely use 
password managers, while prior work has often focused 
on adoption and therefore included participants who do not (yet) use them.

\paragraph{Authenticators:}
Despite receiving fewer reviews (11K), authenticators 
share many similarities with password managers.
Reviewers acknowledged these apps' 
roles in \texttt{Password Protection}: 
keeping their \texttt{Accounts}
and \texttt{Personal Information Protected}. 
Reviewers particularly appreciate using fingerprints 
to approve requests: 
``\textit{\omer{Perfect application, just using your finger print you can 
authenticate apps}}.'' 
In contrast, a minority express concern that their accounts have been compromised despite using the authenticator, or that they cannot access their accounts due to authenticator problems. 
These problems are similar to, but more frequent than, issues reported for password 
managers. Like password managers, authenticators create a single point of authentication failure. Likely due to these issues, 
users expressed much less 
joy (44.2\%) and much more anger (11.7\%) compared to password managers (joy: 67.2\%, anger: 5.6\%).

User concerns about losing account access have also been identified 
in prior work~\cite{owens2021user}. 
Our analysis demonstrates that these concerns are realized, 
some of which could arise from lack of adequate instructional material~\cite{ghorbani2023systematic}. Some work has focused on security 
and privacy weaknesses 
of authenticator implementations~\cite{ozkan2020otp, gilsenan2023otp}. 
However, these weaknesses, by and large, are invisible to users and 
do not appear in reviews.

\paragraph{App Hide and App Lock:}
App hiders attempt to conceal a user-selected list of apps from appearing in the list of installed apps. They generally achieve this functionality by being a \emph{launcher}, 
the default navigation app of the OS. 
App lockers provide access control mechanisms (e.g., a password) before a user is able to open an app, provided the user is using the app locker as the launcher. Although they have related functionality, app lockers do not attempt to hide apps. 

These two \tgtwos together have 148K privacy reviews, of which $\sim$60\% express `Joy.' 
\Themes and \issues overlap almost completely across the two \tgtwos. 
Reviewers of \textit{App Lock} apps praise 
\texttt{Content Hiding} (e.g., 
``\textit{Perfect application. It hides videos, apps, and photos.}'').
Similarly, reviews for \textit{App Hide}
apps express satisfaction with their 
hiding functionality under the \theme \texttt{App Privacy}.

Despite their popularity and users' 
apparent satisfaction with app lockers/hiders, 
surprisingly little research 
has examined these tools. 
Sambasivan et al. found important use 
cases for app lockers among women 
in South Asia~\cite{sambasivan2018privacy}. 
Though we do not have 
gender information, we find that 
India contributed the most reviews for 
these apps. 
Kenya, 
Nigeria, Zambia, Venezuela, 
Pakistan, and \turkey 
also contribute a high rate of these reviews, but 
use in these regions has not, to our knowledge,
been studied. 
Other work points out that device sharing 
creates important threats that are not 
always well supported by developers and 
security professionals~\cite{warford2022sok}. 

Researchers have found some 
app lockers may be 
easily circumventable~\cite{mahmoud2016applock}, 
suggesting that users may be 
less protected than they believe. 
Other research has shown app lockers' resemblance to malware~\cite{bianchi2015androidui, ren2017windowguard}. 
However, despite wide use, there is little 
research analyzing the security, privacy, 
and usability of these 
apps. We argue these apps remain an interesting research topic. 

\paragraph{Photo tools and album management:}
The \textit{Photo tools} \tgtwo has 208K privacy reviews, while \textit{Album management} has 145K.
Reviews for these \tgtwos frequently express `Joy' ($>$45\%), often praising 
\texttt{Photo Protection} (e.g., 
``\textit{I love this app. Photos are ALWAYS protected}'')
and \texttt{Content Hiding} 
(e.g., ``\textit{Greatest application for hiding photos or vids}'')
features. 
Unlike app lockers and hiders, these apps allow users to secure media beyond the initial lock screen.

\paragraph{Diary apps:}
Apps in this 
category receive a perhaps surprising amount 
of privacy-relevant reviews (149K).
These reviews carry an exceptionally positive tone (80.7\% marked with `Joy', the highest rate among all \tgtwos) and 
most often fall under the themes \texttt{Information Privacy} and 
\texttt{Password Protection}. 
Users most often find these apps 
useful to \texttt{Keep secrets} (within the \texttt{Information Privacy} theme):  
``\textit{I really like this. I can keep my secrets. Seriously, thanks.}''
Some users specifically praise the ability 
to password-protect diary 
entries.
These apps are highly regarded and widely used. However, we are unaware of any recent technical analyses of these tools, though some older work exists~\cite{dwan1997security}. 
We recommend future investigations into how secure these apps are.

\paragraph{Antivirus \& task management:}
\textit{Antivirus} \tgtwo has 106K privacy reviews, of which 41\% express `Joy.' \texttt{Password Protection}, 
\texttt{Privacy Concerns}, 
\texttt{App Locking}, 
\texttt{Personal Information Privacy}, 
and \texttt{Data Deletion} 
are the most common \themes. 
Users comment about a variety of issues outside the traditional antivirus app functionality, which we posit relates to antivirus apps expanding their feature offerings\begin{usenix}
.\footnote{\url{https://www.pcmag.com/picks/the-best-security-suites}}
\end{usenix}
\begin{arxiv}
\cite{rubenking2023securitysuitelist}. 
\end{arxiv}
For example, we see praise for app locking features not only in apps categorized as app lockers, but also in general antivirus apps, many of which offer such features.
We see relatively fewer mentions of malware (the primary purpose of antivirus software); examples from
\issues like \texttt{Phone Protection} and \texttt{Malware Protection}, include: 
``\textit{This is a perfect app. It finds and removes malware very fast. Further, it detects apps that are a threat to your privacy. I 
100\% recommend it.}''

The \textit{Task \& app management} \tgtwo has 51k privacy reviews, of which 35\% are privacy positive.
These apps
exhibit similar trends to antivirus apps, 
perhaps because of large overlap in functionality: both managers and antivirus programs offer security and performance 
features. 
However, reviews of task management apps include more reviews about  \texttt{Excessive permissions}. Though both \tgtwos require sensitive permissions to function (e.g., 
accessing all files), task \& app 
management users may be more hesitant 
to grant these permissions outside a security context: 
``\textit{what is up with these crazy permissions? thanks. nope.}''

\takeaways Within the 32\% of privacy reviews that are positive, users praise privacy-protecting mechanisms (data deletion, hiding, password-protected access) for several data types (photos, diary entries, app visibility, browsing history).

\section{Discussion \& Conclusion}

In examining more than a decade of Google Play app reviews, we find that privacy reviews are growing at a biannual rate of 9\%. \texttt{Data Deletion} is the number one privacy issue worldwide in terms of total reviews and is a top-5 concern in 47 countries. While this issue is growing, other prevalent issues such as those around permissions (e.g., \texttt{Excessive Permissions}) are declining. We further illustrated how an emotions classifier can illuminate which app types cause the most privacy concern and which receive the most privacy praise, across an entire app store.

User research on \sp tools rarely focuses on what users appreciate; rather, it highlights 
shortcomings (e.g., ~\cite{lee2017usability, ramesh2022vpnconsumer, story2021awareness, huaman2021pm}). Our analysis adds perspective
by unearthing several potential {\bf privacy wins}.
First, we found a large body of positive privacy reviews for app lockers and hiders, 
VPNs, journaling apps, 
and album management. 
Many of these app categories are understudied; 
future work could investigate them in more depth, 
including improvements to support user workflows and technical analyses of whether 
apps are providing the privacy users expect. Second, we see that, over time, the ratio of permission-related complaints in privacy reviews has dropped by more than half (\autoref{sec:permissions}). This steep decline suggests permissions concerns could be abating, after concerted effort to improve permissions systems. This hypothesis would be well served by a future user study.

Our analysis can also be leveraged to \textbf{identify anomalous behaviors}, such as countries with unique patterns of topic discussions (e.g., Nigeria and \turkey), or countries with abnormally high privacy-review growth rates (e.g., Indonesia). 
We uncovered large groups of reviewers who use reviews to communicate with developers in ways unlikely to achieve their goals~\cite{google2023policy}. Examples include ineffective quasi-legal privacy disclaimers in \turkey, as well as data deletion requests for specific accounts in Indonesia.

As previously noted~\cite{harkous20222hark}, this privacy-review analysis could {\bf help developers}. Broadly speaking, we see two types of privacy feedback. Non-specific feedback, such as reviews that say ``\textit{this is privacy invasive, do not install},'' are not directly actionable, but they do offer developers an understanding of privacy sentiment, and their volume matters since they often discourage other users from installing an app. More specific feedback can be turned into actionable insights (e.g., ``\textit{please add private chat}'').
Finally, privacy-positive reviews may offer developers a new kind of privacy success metric.

We argue that our analysis approach can be useful for a variety of {\bf research purposes}. We have shown in this paper examples of: (1) 
\textit{corroborating existing research}, such as the broad public
support for VPNs~\cite{dutkowska2022vpn} (in~\autoref{sec:securityprivacy}), and the divergent views about child monitoring apps~\cite{cranor2014parents,wei2022anti}; (2) \textit{adding context to prior work}, such as showing that privacy concerns about financial loan apps from Kenya ~\cite{munyendo2022kenya} are being similarly reported in Indonesia, India, and Thailand (in ~\autoref{sec:loan_apps}); and (3) \textit{identifying unexpected or 
emerging privacy issues} that can seed future work using interviews (to enable direct discussion with users) or surveys (that can direct users to focus on specific aspects). Examples include examining if diary apps and app lockers actually deliver the privacy features offered, and whether users correctly understand the privacy offerings of those apps, in addition to exploring why permissions are a decreasing topic of discussion.

Recent advances in large language models (LLMs), such as OpenAI's GPT-4~\cite{gptfour2023} or \omer{Google's Gemini~\cite{gemini},} 
have shown \omer{promising performance gains on 
multiple natural language tasks.} 
However, these LLMs \omer{require significant prompt engineering (see~\autoref{app:classifier_baselines}),} 
have high inference costs, and cannot easily scale to large datasets, unless they have been distilled to smaller models~\cite{hinton2015distilling}. Exploring such 
distillation approaches for the purpose of 
improving the models used in this work is a 
natural avenue of future work. 

In summary, we show that large-scale text analysis, 
followed by zooming in to explore changing trends or unusual 
patterns, is useful to both summarize the privacy pulse as it 
ebbs and flows across much of the globe, as well as to 
surface privacy issues that may not be regularly tracked.

\section{Acknoledgements}
We would like to thank Kurt Thomas and Patrick Gage Kelley 
for their insightful comments on earlier manuscripts. We also thank the anonymous 
USENIX reviewers for their feedback, and the Google Play team for providing data access and removing spying reviews we identified.

\bibliographystyle{plain}
\bibliography{refs}

\begin{arxiv}
    \clearpage
\end{arxiv}

\begin{center}
    {\bf \large Appendix}
\end{center}

\appendix

\section{Translated Languages}
\label{sec:translated_languages}
These 24 languages were translated (percentage of our data):
Spanish (13.3\%), Indonesian (9.3\%), Portuguese (7.8\%), Russian (5.8\%), Turkish (4.2\%), German (3.3\%), Arabic (2.5\%), French (2.3\%), Italian (1.7\%), Korean (1.7\%), Vietnamese (1.1\%), Polish (1\%), Persian (0.6\%), Thai (0.5\%), Dutch (0.5\%), Romanian (0.2\%), Czech (0.2\%), Hungarian (0.2\%), Ukrainian (0.1\%), Greek (0.1\%), Chinese (0.1\%), 
Japanese (0.1\%), Malay (0.07\%), Hindi (0.06\%).

\sai{
\section{Privacy Classifier Baselines}
\label{app:classifier_baselines}

We investigated various modern transformer architectures 
(excluding traditional models, such as BiLSTM, which are inferior), 
and selected six models based on comparisons made within the recently published and extensively cited DeBERTaV3 paper~\cite{he2021debertav3}. In addition, we also considered BART, T5, and FLAN-T5 models for their popularity. 
We chose the largest available variant of DeBERTaV3, and chose other models’ sizes accordingly. 
We use Hark's~\cite{harkous20222hark} T5-11B model as a comparative baseline; our T5-11B model only differs in the diverse training dataset used (\autoref{sec:hark_modifications}). 
This list of models is not exhaustive, but we believe it offers a reasonable baseline.

All models were trained and tested on the datasets created in~\autoref{sec:hark_modifications}. Each model was trained with parameter optimization and `early stopping' (using validation loss as the metric) to avoid overfitting. Each model's best trained variant (highest obtained F1) are reported in \autoref{tab:privacy_classifier_baselines}. Our findings suggest that our T5-11B model delivers the best performance across the board, with a highest ROC-AUC of 0.95 and F1 of 0.87. 

In addition, we also experimented with Gemini 1.0 Pro,
a state-of-the-art (closed-source) LLM~\cite{gemini}. 
Despite our prompt engineering (i.e., trying multiple prompt variants and using few-shot examples), 
we could not better the performance T5-11B. 
Our prompts described our privacy taxonomy, where each high level concept was defined using fine-grained aspects. 
The prompts further included instances of reviews and their 
labels as few-shot examples. For the best performing prompt}
\begin{usenix}
(see the extended paper~\cite{akgul2024reviews}),
\end{usenix}
\begin{arxiv}
(see~\autoref{app:prompt}),
\end{arxiv}
the Gemini Pro performed well on recall (0.83) but precision suffered (0.77), with a ROC-AUC of 0.87. We hypothesize that the poor precision arises because the privacy nuances captured in our carefully annotated training dataset are hard to establish using constrained prompts. Exploring LoRA techniques~\cite{hu2021lora} to finetune an LLM for this task is outside the scope of this paper.

\begin{table}[ht]
    \small
    \begin{center}
        \begin{tabular}{{m{2.4cm}|m{0.70cm}|m{0.45cm}|m{0.45cm}|m{0.45cm}|c}}
        \toprule
        \hline
        Model & Accur. & F1 & P & R & ROC-AUC \\
        \hline
        \hline
        $Ours\ (T5_{11B})$ & \g{0.89}	& \g{0.87} & \g{0.91} & \g{0.84} & \g{0.95} \\
        \hline
        $DeBERTaV3_{Large}$ & \g{0.85} & \g{0.84} & \g{0.81} & \g{0.87} & \g{0.92} \\
        \hline
        $T5_{Large}$ & \g{0.83} & \g{0.83} & \g{0.77} & \g{0.90} & \g{0.92} \\
        \hline
        $ELECTRA_{Large}$ & \g{0.85} & \g{0.82} & \g{0.88} & \g{0.76} & \g{0.90} \\
        \hline
        $FLAN-T5_{Large}$ & \g{0.84} & \g{0.81} & \g{0.85} & \g{0.77} & \g{0.91} \\
        \hline
        $RoBERTa_{Large}$ & \g{0.82} & \g{0.81} & \g{0.79} & \g{0.82} & \g{0.88} \\
        \hline
        $Gemini_{Pro}$ & \g{0.82} & \g{0.80} & \g{0.77} & \g{0.83} & \g{0.87} \\
        \hline        
        $BART_{Large}$ & \g{0.81} & \g{0.80} & \g{0.76} & \g{0.84} & \g{0.88} \\
        \hline
        $Hark\ (T5_{11B})$ & \g{0.79} & \g{0.77} & \g{0.73} & \g{0.81} & \g{0.87} \\
        \hline
        $ALBERT_{Large}$ & \g{0.79} & \g{0.76} & \g{0.76} & \g{0.75} & \g{0.84} \\
        \hline
        $BERT_{Large}$ & \g{0.78} & \g{0.75} & \g{0.74} & \g{0.77} & \g{0.84} \\
        \hline
        $XLNet_{Large}$ & \g{0.74} & \g{0.72} & \g{0.67} & \g{0.79} & \g{0.82} \\
        \hline
        \bottomrule
        \end{tabular}
    \caption{Privacy classifier performance. P: precision, R: recall.}
    \label{tab:privacy_classifier_baselines}

    \vspace{20cm}

    \end{center}
\end{table}

\begin{arxiv}
\begin{figure*}
    \section{Heatmap Showing Theme Distribution per Country.}
    \label{app:cluster}
    \centering
    \includegraphics[width=.95\textwidth]{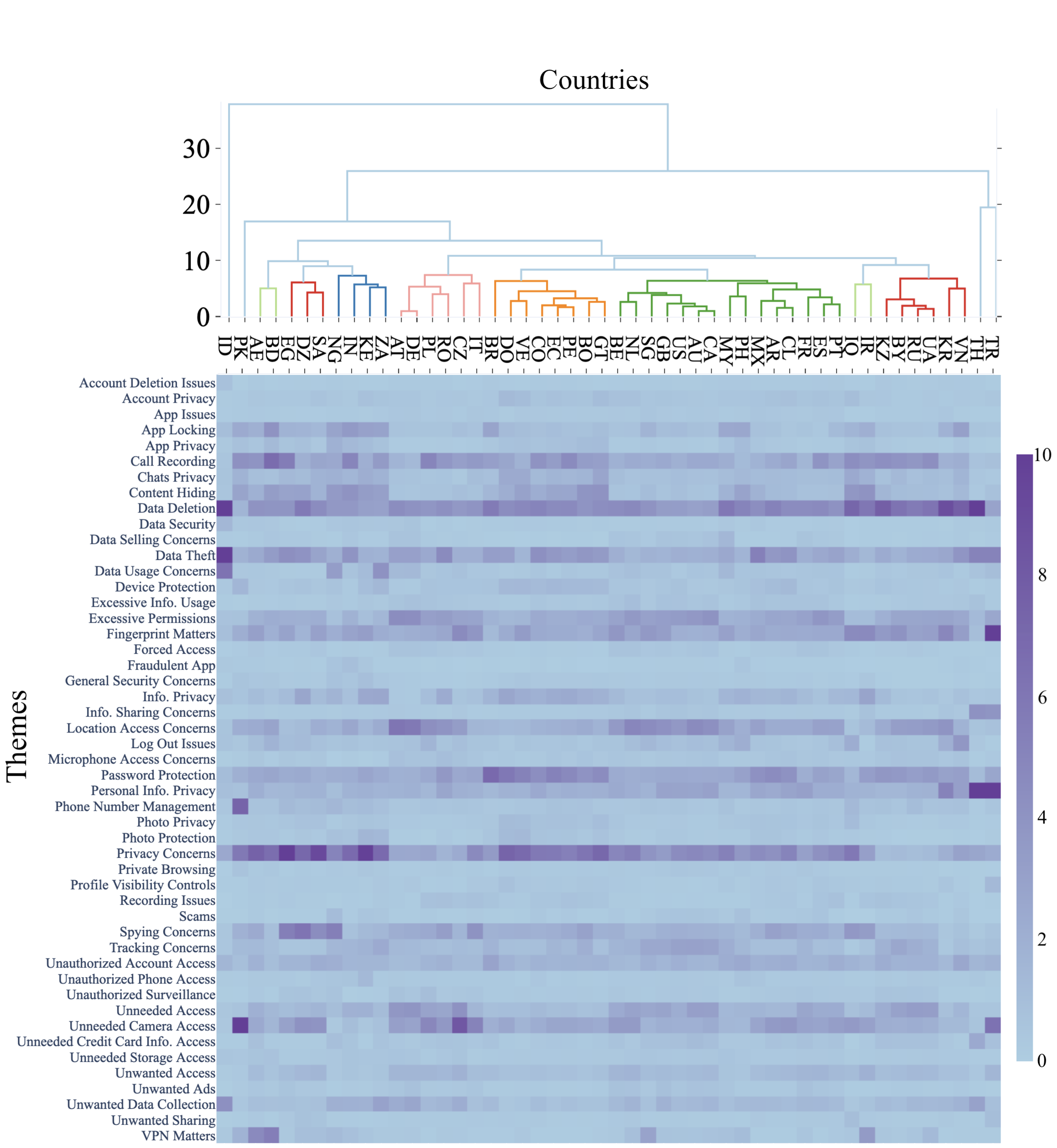}
    \caption{(Top) Dendrogram of hierarchical clustering of countries based vectors visualized below. (Bottom) Percentage of a \theme within the privacy reviews from each country. Darkest color indicates 10\% or more. Lightest color indicates 0\%. Both graphs share the same order of countries (x axis).}
    \label{fig:cluster}
\end{figure*}
\end{arxiv}

\begin{arxiv}

\begin{table*}[ht!]
\section{Expanded Privacy Taxonomy}
\label{app:taxonomy}
   \footnotesize
    \centering
    \begin{tabular}{cc}
    \begin{tabular}{|c|l|}
    \hline
    Concept & Fine-grained aspect \\
    \hline
    \hline
    \multirow{5}{6em}{Data Collection} & Collecting unnecessary personal data \\
    \cline{2-2}
    & Purpose of data collection \\
    \cline{2-2}
    
    \cline{2-2}
    & Data Aggregation\\
    \cline{2-2}
    & Data Minimization \\
    \hline
    
    \multirow{7}{6em}{Data Sharing} & Accidental sharing \\
    \cline{2-2}
    & Forced sharing \\
    \cline{2-2}
    & Unintended sharing \\
    \cline{2-2}
    & Cross-app sharing \\
    \cline{2-2}
    & Secondary use \\
    \cline{2-2}
    & Calendar sharing \\
    \cline{2-2}
    & Password protected sharing\\
    \hline
    
    \multirow{9}{6em}{Data Deletion} & Browsing history removal \\
    \cline{2-2}
	& Delete Video history \\
	\cline{2-2}
	& Delete search history \\
	\cline{2-2}
	& Cookie removal \\
	\cline{2-2}
	& Photo deletion \\
	\cline{2-2}
	& Video deletion \\
	\cline{2-2}
	& File deletion \\
	\cline{2-2}
	& Profile deletion \\
    \hline
    
    \multirow{5}{6em}{Remove Personally Identifiable Information} & Remove name \\
    \cline{2-2}
	& Remove address \\
	\cline{2-2}
	& Remove phone number \\
	\cline{2-2}
	& Remove Personal information \\
	\cline{2-2}
	& Remove photo \\
	\hline
	
	\multirow{6}{6em}{Data Exposure} & To advertisers \\
	\cline{2-2}
	& To app developers \\
	\cline{2-2}
	& Public accessibility \\
	\cline{2-2}
	& Data disclosure \\
	\cline{2-2}
	& Safety \\
	\cline{2-2}
	& Use Limitation \\
	\hline
	
	\multirow{13}{6em}{Data Hiding}	& Name \\
	\cline{2-2}
	& Video \\
	\cline{2-2}
	& Events \\
	\cline{2-2}
	& Albums \\
	\cline{2-2}
	& Photos \\
	\cline{2-2}
	& Contacts \\
	\cline{2-2}
    & Display name \\
    \cline{2-2}
	& Email \\
	\cline{2-2}
	& Messages \\
	\cline{2-2}
	& Notes \\
	\cline{2-2}
	& Playlist \\
	\cline{2-2}
	& Folders \\
	\cline{2-2}
	& Meetings \\
	\hline
	
	\multirow{2}{6em}{Location and Tracking} & Location data \\
	\cline{2-2}
	& Tracking \\
	\hline

    \end{tabular} & 
    
    \begin{tabular}{|c|l|}
    \hline
    Concept & Fine-grained aspect \\
    \hline
    \hline
    \multirow{8}{6em}{Consent} & Giving consent \\
    \cline{2-2}
	& Agreement \\
	\cline{2-2}
	& Authorize \\
	\cline{2-2}
	& Consent process \\
	\cline{2-2}
	& Opt-out (notice awareness) \\
	\cline{2-2}
	& Without consent \\
	\cline{2-2}
	& Mis-activation \\
	\cline{2-2}
	& Forced consent \\
	\hline
	
	\multirow{7}{6em}{Privacy Controls} & Privacy settings \\
	\cline{2-2}
	& Changing personal information \\
	\cline{2-2}
	& Location sharing \\
	\cline{2-2}
	& Download my data \\
	\cline{2-2}
	& Parental controls \\
	\cline{2-2}
	& Privacy defaults \\
	\cline{2-2}
	& Password protected controls \\
	\hline
		
	\multirow{5}{6em}{Anonymity / Identification} & User wants to be anonymous \\
	\cline{2-2}
	& Incognito \\
	\cline{2-2}
	& Hide group participants \\
	\cline{2-2}
	& Fear of identification \\
	\cline{2-2}
	& Misattribution \\
    \hline
    
    \multirow{2}{6em}{Advertising}	& Ads personalization \\
    \cline{2-2}
	& Paid services\\
	\hline
	
	\multirow{6}{6em}{Data Security} & Safety \\
	\cline{2-2}
	& Protection \\
	\cline{2-2}
	& Breach of confidentiality \\
	\cline{2-2}
	& Data breach \\
	\cline{2-2}
	& Account hacking \\
	\cline{2-2}
	& Fake profiles \\
	\hline	
	
	Password Issues &  - \\
	\hline
	
	\multirow{3}{6em}{Data Accuracy} & Inaccurate \\
	\cline{2-2}
	& Obsolete \\
	\cline{2-2}
	& Wrong association	\\
    \hline
    
    \multirow{4}{6em}{Safety}	& Blackmail \\
    \cline{2-2}
	& Appropriation \\
	\cline{2-2}
	& Intrusion \\
	\cline{2-2}
	& Decisional interference \\
	\hline
	
	\multirow{2}{6em}{Selling data} & To 3rd parties \\
	\cline{2-2}
	& For ads purposes \\
	\hline
	
	\multirow{3}{6em}{Surveillance} & Spying \\
	\cline{2-2}
	& Stalkerware \\
	\cline{2-2}
 	& Satellite \\ 
	\hline

	Privacy Invasion & - \\
	\hline
    Privacy policies and laws & - \\
    \hline
    Positive Privacy & - \\

	\hline
    \end{tabular} \\
    \end{tabular}
    \caption{Privacy taxonomy. Takes~\cite{harkous20222hark} (derived from academic taxonomies) 
    and expands per material found in reviews (expert manual analysis). 
    Note this does not represent any formal corporate taxonomy.}
    \label{tab:expanded_privacy_taxonomy}
\end{table*}

\end{arxiv}

\begin{arxiv}
\newpage
\begin{table*}
    \section{Privacy Classifier Prompt (derived from~\autoref{app:taxonomy})}
    \label{app:prompt}
    \begin{adjustbox}{width=1.16\textwidth,center}
    \begin{tabular}{p{7in}}
    \fontsize{5}{7}\selectfont

    You are a world-leading expert in data labeling.
    
    We want you to label each "Input" text as being related to Privacy. A text is considered to be related to Privacy, if it explicitly mentions any of the subconcepts defined below:
    
    - Data Collection: Data collection is a privacy issue when it relates to collecting personal data from apps and services. If online apps and services collect personal data that is viewed as unneeded or excessive for the functionality of the service, then it is a privacy issue. Other examples of data collection issues that relate to privacy include: when the purpose of data collection is not shared, when the purpose of data collection is poorly explained, when data about a user is collected from external sources other than the direct app or web service they are using, when more data is gathered than the minimum needed for the app functionality, or when forced to link accounts or link different types of personal information.
    
    - Data Sharing: Data sharing is a privacy issue and refers to scenarios when users want sharing controls to manage which friends, family, or social media see their user generated content; user generated content includes photos, videos, calendars, contacts, passwords and files. Users also want data sharing control manage the sharing of user behavior data, such as location or activity tracking, according to their privacy preferences. Negative examples of data sharing issues that relate to privacy include: 1) when users feel forced to share their personal content with more people or services than they would like, or 2) when oversharing happens (such as unintended or accidental sharing)  due to confusion with sharing controls, or 3) when data is shared across multiple apps or services without explicit user consent. Privacy positive examples include 1) when a user is pleased with a privacy control, or 2) pleased about the functionality of a privacy feature.
    
    - Data Deletion: Data deletion is a privacy issue when the user wants to delete data for a privacy reason; this includes deleting data that is considered personal or deleting data the user is afraid is visible by others in cases when they did not consent. If the user is having trouble deleting their personal data because they cannot find the deletion controls then it is a privacy issue. Similarly, if the privacy controls for data deletion do not have the intended affect, then it is a privacy concern.  Personal data that a user might want to delete for privacy reasons includes user generated data (such as files, images, videos, messages, emails and profile data) as well behavior history (such as online searching, youtube watching history, location tracking history, web cookies, and website browsing data).
    
    - Remove Personally Identifiable Information: Removing Personally Identifiable Information (PII) is a privacy issue when a user wants their name, address, phone number, birthdate, personal photo or profile info removed from a public place. It is a privacy issue when users say they want their PII to be private or non-public, or if they feel visibility of this information was something they did not consent to. However it is not a privacy issue when the reason a user wants to remove PII is because it is inaccurate.
    
    - Data Exposure: Data Exposure is a privacy issue when 1) a user does not have direct control over their personal data, and it may be shared with 1st parties, 3rd parties or advertisers; 2) when users do not have fine-grained privacy controls and cannot influence third party sharing on a per session basis; 3)  when users think that exposure of their personal data poses a safety threat; and 4) when users are upset that their personal data has been shared with the public at large. This differs from Data Sharing that typically involves a user explicitly deciding on a per item basis (files, videos, photos) who to share with.
    
    - Data Hiding: When users want to hide their personal data it is a privacy issue. Examples of data hiding that relate to privacy include wanting the needed controls to hide from others data such as name, contacts, photos, videos, personal events, email, messages, notes, playlists, folders, meetings and profile data. Note that data deletion of an account is not considered a data hiding issue. When users ask how to work the controls, without mentioning privacy or hiding personal data, then it is not a privacy issue.
    
    - Location and Tracking: Location tracking refers to technology that tracks the location and movements of people. Because location tracking data is considered private, the following examples are all considered privacy issues: 1) when users write about wanting to delete their location history; 2) when users want control over who sees their location data; 3) want to know who has access to their location data; 4) when users want location tracking to be turned off or feel location controls are not working as expected.
    
    - Surveillance: Digital surveillance that occurs when apps or web services are used for spying or stalking on people is a privacy issue. Digital spying refers to  the continuous or repeated 1) listening to people with the help of technical devices  such as microphones for recording voices, 2) viewing, photographing or videotaping of individuals via cameras or satellites, or 3) collection of a persons location. When such ongoing monitoring data is given to or collected from an unexpected person or agency, it is a privacy problem. Examples include 1) the use of spyware or stalkerware apps, 2) when cameras or microphones are turned on and the user is unaware that they are on, and 3) when such data is shared with governments or public agencies.
    
    - Consent: When users write about giving consent, or authorization, or approval, to an app or web service to collect their personal data it is a privacy issue. Consent is also a privacy issues when users feel they have not had the opportunity to consent to the collection or sharing of their personal data. When users discuss consent in the context of privacy, they may say 1) they were or were not able to give consent easily, or 2) that something happened without having given consent (like they observed data collection or data sharing), or 3) they may discuss an opt-in or opt-out option related to data sharing. When an app or service activates without prompting and a user thinks they did not consent to this activation, then it is a privacy issue.
    
    - Privacy Controls: When a user discusses any aspect of privacy controls then it is a privacy issues. Privacy controls are commonly provided for microphones, cameras, location tracking, and web browsing. Providing additional passwords to manage which friends and family members see user created content such as files, albums, images, videos, emails and chats, is another type of privacy control. Parental controls are used to manage the privacy settings of their children's accounts.  For such privacy controls, users expect to  1) know how to change them, 2) feel they should be easy to use, 3) to understand what the default settings are,  5) to understand how to download one's private data. When a user feels their privacy choices and preferences are not being honored, then it is a privacy issue.
    
    - Anonymity / Identification: When a user discusses using digital apps and web services in an anonymous fashion, without identifying themselves, then it is a privacy issue. Examples include when a user wants to browse anonymously or in incognito mode, seeks to understand if anonymous borrowing really works, what level of anonymity is provided, or if partial data may be collected when browsing anonymously. Users may want to be anonymous so that websites cannot carry out fingerprinting to identify them. Another privacy aspect of anonymity is misattribution - if users discuss being mistakenly assumed to be someone else, then a privacy issue has occurred.
    
    - Selling data: When one organization or party sells a users private data to a different organization, it is a privacy issues. 1st party services may sell personal data to 3rd party services for the purposes of advertising. This should require the user's consent.
    
    - Advertising: When personal data is used for advertising, it is a privacy issue. It is a privacy issue when 1) users try turn off ads personalization but nevertheless still fear their online activities are being tracked, as this indicates their privacy preferences are not being honored; 2) users would rather pay to stop ads in order to protect their private data from being used for advertising.  When ads are very personal or closely related to a recent user activity, users can feel spooked that their online behavior is being observed and used.
    
    - Data Security: Data security is a privacy issue when 1) users are concerned about the safety and protection of their data, or 2) if a data breach is discussed. Account hacking and safety are a privacy issue when the user discusses 1) that the effect of hacking was to steal or exposure personally identifying information (PII), or 2) that their account is being used by someone else without consent such as in the case of impersonation or via the use of fake profiles.
    
    - Password Issues: Password issues can sometimes be a privacy issue. One example is when someone writes about wanting password protection as a means to control sharing of personal data, such as files, videos, photos; password protection controls sharing for privacy reasons by only allowing people who have been given the password to see the personal data. Another example is when a user gives someone else (friend, family, colleague) a password and that other person abuses the password to impersonate the user, or take over control of their devices, or reset the original password so the user loses control of their device and account.
    
    - Data Accuracy : Data accuracy is a privacy issue when personal data is inaccurate (such as in an address) or obsolete (no longer valid); these are privacy issues because inaccuracies in personal information can lead to misattribution.
    
    - Safety: Safety issues that occur due to personal data exposure or exploitation
    
    - Privacy Invasion: When a user writes that they feel a violation of their privacy, or that their privacy has been invaded, it is a privacy issue.
    
    - Privacy policies and laws: If a user writes anything about a privacy policy, like where it is or it being hard to understand, then it is a privacy issue. If a user writes anything about privacy laws, or consumer data protection acts, or personal data protection bills, it is a privacy issue. Examples of well known privacy laws and regulations include  GDPR, CCPA, CPRA, COPPA, and HIPPA for privacy legal protection of health data.
    
    - Positive Privacy: Privacy issues can contain positive sentiment. Examples of positive privacy issues include: when users are happy with privacy controls, or data protection, or data deletion controls,  or specific privacy features offered in a product.

    A text is NOT related to Privacy:
    - If it is discussing a feature or functionality not working as expected, and does not explicitly any of the privacy subconcepts earlier.
    
    - If it is complaining that the app consumes or takes too much data, which could be related to mobile data bandwidth usage.
    
    Texts that are vague and that do not explicitly mention any of the privacy subconcepts should be labeled as 'not-privacy'.
    
    In addition to predicting a 'privacy' or 'not-privacy' label, do share your reasons for the predicted label.
    
    Here are a few examples that explain the task:
    
    Input: This app keeps crashing. I have restarted my phone, re-installed the app. Nothing works. I am so angry.
    
    Answer ('privacy' or 'not-privacy'): not-privacy
    
    Reason: The text pertains to the app not working, and does not relate to any of the privacy subconcepts defined.
    
    Input: I want to hide my transaction history. Why am I not able to hide channels that I don't want to watch.
    
    Answer ('privacy' or 'not-privacy'): privacy
    
    Reason: The text discusses hiding transaction history and channels, which relates to 'Data Exposure' privacy subconcept.
    
    Input: This app does not properly save my data. I have lost a lot of my data. I am so angry.
    
    Answer ('privacy' or 'not-privacy'): not-privacy
    
    Reason: The text pertains to the app not properly saving user's data. It is complaining about the app not working properly,and does not relate to any of the privacy subconcepts defined.
    
    Input: You need a picture of my driver's license to send me targeted advertising.
    
    Answer ('privacy' or 'not-privacy'): privacy
    
    Reason: The text discusses a user complaint about sharing driver's license information, which relates to 'Data Collection' privacy subconcept.
    
    Input: This app is taking too much data.
    
    Answer ('privacy' or 'not-privacy'): not-privacy
    
    Reason: The text pertains to the app taking too much data. As the text is vague, it could be about the app consuming more mobile bandwidth data, and it may not relate to any of the privacy subconcepts defined.
    
    Input: Do not let me log in...
    
    Answer ('privacy' or 'not-privacy'): not-privacy
    
    Reason: The text pertains to user suggesting/requesting the app not to let them log in. As the text is vague, it could be due to a functional issues and is not related to any of the privacy subconcepts defined.
    
    Based on the Privacy definition above, is the below input text related to the concept of Privacy.
    
    Input: {{input}}
    Answer ('privacy' or 'not-privacy'): \\
\end{tabular}
\end{adjustbox}
\end{table*}

\end{arxiv}

\end{document}